\documentclass[twoside,11pt]{article}
\usepackage{jair, theapa, rawfonts}

\usepackage[colorinlistoftodos]{todonotes}

\usepackage{graphicx}
\usepackage{epstopdf}
\usepackage{caption}
\usepackage{subcaption}
\usepackage{wrapfig}

\usepackage{enumitem}

\usepackage{tikz}
\usepackage{tabularx}
\usepackage{placeins}
\usepackage[hang,flushmargin]{footmisc}   %

\usepackage{algorithm}
\usepackage{algpseudocode}
\usepackage{pifont}

\usepackage{amsmath}
\usepackage{amssymb}
\usepackage{amsfonts}
\usepackage{amsthm}
\usepackage{mathtools}
\usepackage{bm}
\usepackage{mathrsfs}

\usepackage{setspace}
\usepackage{xspace}
\newcommand{\etal}{et~al.\xspace}

\newcommand{\Task}[1]{\mathcal{Z}^{(#1)}}
\newcommand{\Mt}[1]{\phi\bigl(\mt{#1}\bigr)}
\newcommand{\mt}[1]{\bm{m}^{(#1)}}

\newcommand{\st}[1]{\bm{s}^{(#1)}}

\newcommand{\Th}[1]{\bm{\theta}^{(#1)}}
\newcommand{\At}[1]{\bm{\alpha}^{(#1)}}
\newcommand{\ComboVec}[1]{\bm{\beta}^{(#1)}}
\newcommand{\ComboDict}{\bm{K}}

\newcommand{\Xt}[1]{\bm{X}^{(#1)}}

\newcommand{\yt}[1]{\bm{y}^{(#1)}}

\newcommand{\Loss}{\mathcal{L}}

\newcommand{\Reals}{\ensuremath{\mathbb{R}}}

\newcommand{\comE}[1]{\textcolor{blue}{\fontfamily{cmtt}\selectfont \textbf{\large [\ding{42}}\colorbox{yellow}{\textbf{Eric:}} {\small #1}\textbf{\large ]}}}

\newtheorem{theorem}{Theorem}[section]

\newcommand{\transpose}{\text{$\mathsf{T}$}}

\newcommand{\thetaprime}{\tilde{\bm{\theta}}}

\newcommand{\objJtheta}{\mathcal{J\!}_{\!\mathcal{L},\bm{\theta}}}
\newcommand{\trajectories}{\mathbb{T}}
\newcommand*\diff{\mathop{}\!\mathrm{d}}
\newcommand{\taut}{\bm{\tau}}

\newcommand\blfootnote[1]{%
  \begingroup
  \renewcommand\thefootnote{}\footnote{#1}%
  \addtocounter{footnote}{-1}%
  \endgroup
}

\newcommand{\comment}[1]{}

\jairheading{?}{2017}{1-15}{2/17}{?/??}
\ShortHeadings{Using Task Descriptions in Lifelong Learning}
{Isele, Rostami, \& Eaton}
\firstpageno{1}

\begin{document}

\title{Using Task Descriptions in Lifelong Machine Learning\\for Improved Performance and Zero-Shot Transfer}
\author{\name David Isele \email isele@seas.upenn.edu \\
       \name Mohammad Rostami \email mrostami@seas.upenn.edu \\
       \name Eric Eaton \email eeaton@seas.upenn.edu \\
       \addr University of Pennsylvania,\\
       Philadelphia, PA  19104 USA}

\maketitle

\begin{abstract}
Knowledge transfer between tasks can improve the performance of learned models, but requires an accurate estimate of the inter-task relationships to identify the relevant knowledge to transfer. These inter-task relationships are typically estimated based on training data for each task, which is inefficient in lifelong learning settings where the goal is to learn each consecutive task rapidly from as little data as possible. To reduce this burden, we develop a lifelong learning method based on coupled dictionary learning that utilizes high-level task descriptions to model the inter-task relationships. We show that using task descriptors improves the performance of the learned task policies, providing both theoretical justification for the benefit and empirical demonstration of the improvement across a variety of learning problems. Given only the descriptor for a new task, the lifelong learner is also able to accurately predict a model for the new task through zero-shot learning using the coupled dictionary, eliminating the need to gather training data before addressing the task.
\end{abstract}

\section{Introduction}
\label{Introduction}

Transfer learning (TL) and multi-task learning (MTL) methods reduce the amount of experience needed to train individual task models by reusing knowledge from other related tasks. This transferred knowledge can improve the training speed and model performance, as compared to learning the tasks in isolation following the classical machine learning pipeline \cite{Pan2010a}.  TL and MTL techniques typically select the relevant knowledge to transfer by modeling inter-task relationships using a shared representation, based on training data for each task  \cite{baxter2000model,ando2005framework,Bickel2008,Maurer2013}.  
Despite benefits over single-task learning, this process  requires sufficient training data for each task to identify these relationships before knowledge transfer can succeed and improve generalization performance.  This need for data is especially problematic in learning systems that are expected to rapidly learn to handle new tasks during real-time interaction with the environment: when faced with a new task, the learner would first need to gather data on the new task before bootstrapping a model via transfer, consequently delaying how quickly the learner could address the new task.  \blfootnote{$^\dagger$An earlier version of this work focusing on policy gradient reinforcement learning appeared in the proceedings of IJCAI 2016 \cite{isele2016task}.} 

Consider instead the human ability to rapidly bootstrap a model for a new task, given {\em only} a {\em high-level task description}---before obtaining experience on the actual task.  For example, viewing only the image on the box of a new IKEA chair, we can immediately identify previous related assembly tasks and begin formulating a plan to assemble the chair. In the same manner, an experienced inverted pole balancing agent may be able to predict the controller for a new pole given its mass and length, prior to interacting with the physical system.  These examples suggest that an agent could similarly use high-level task information to bootstrap a model for a new task more efficiently.

Inspired by this idea, we explore the use of high-level task descriptions to improve knowledge transfer between multiple machine learning tasks.  We focus on lifelong learning scenarios~\cite{Thrun1996,Ruvolo2013}, in which multiple tasks arrive consecutively and the goal is to rapidly learn each new task by building upon previous knowledge. Our approach to integrating task descriptors into lifelong machine learning is general, as demonstrated on applications to reinforcement learning, regression, and classification problems. 

Our algorithm, Task Descriptors for Lifelong Learning (TaDeLL), encodes task descriptions as feature vectors that identify each task, treating these descriptors as side information in addition to training data on the individual tasks.  The idea of using task features for knowledge transfer has been explored previously by Bonilla~\etal~\citeyear{bonilla2007kernel} in an offline batch MTL setting, and more recently by Sinapov~\etal~\citeyear{Sinapov2015} in a computationally expensive method for estimating transfer relationships between pairs of tasks.  In comparison, our approach operates online over consecutive tasks and is much more computationally efficient.

We use {\em coupled dictionary learning} to model the inter-task relationships between the task descriptions and the individual task policies in lifelong learning.  
The coupled dictionary enforces the notion that tasks with similar descriptions should have similar policies, but still allows dictionary elements the freedom to accurately represent the different task policies. We connect the coupled dictionaries to the concept\comment{s of co-training \cite{blum1998combining} and} of mutual coherence in sparse coding, providing theoretical justification for why the task descriptors improve performance, and verify this improvement empirically.

In addition to improving the task models, we show that the task descriptors enable the learner to accurately predict the policies for unseen tasks given only their description---this process of learning without data is known as {\em zero-shot learning}.  This capability is particularly important in the online setting of lifelong learning. It enables the system to accurately predict policies for new tasks through transfer, without requiring the system to pause to gather training data on each task. 

Specifically, this article provides the following contributions:
\vspace{-0.5\topsep}
\begin{itemize}[itemsep=0pt]
\item We develop a general mechanism based on {\bf coupled dictionary learning} to incorporate task descriptors into into knowledge transfer algorithms that use a factorized representation of the learned knowledge to facilitate transfer \cite{Kumar2012,Maurer2013,Ruvolo2013}.  
\item Using this mechanism, we develop {\bf two algorithms}, for lifelong learning (TaDeLL) and MTL (TaDeMTL), that incorporate task descriptors to improve learning performance.  %
\item Most critically, we show how these algorithms can achieve {\bf zero-shot transfer} to bootstrap a model for a novel task, given only the high level task descriptor.
\item We provide {\bf theoretical justification} for the benefit of using task descriptors in lifelong learning and MTL, building on the idea of mutual coherence in\comment{co-training and} sparse coding.
\item Finally, we demonstrate the empirical effectiveness of TaDeLL and TaDeMTL on {\bf reinforcement learning} scenarios involving the control of dynamical systems, and on prediction tasks in {\bf classification and regression} settings, showing the generality of our approach. 
\end{itemize}

\section{Related Work} \label{sect:RelatedWork}

Multi-task learning (MTL) \cite{Caruana1997} methods often model the relationships between tasks to identify similarities between their datasets or underlying models. There are many different approaches to modeling these task relationships. Bayesian approaches take a variety of forms, making use of common priors~\cite{wilson2007multi,lazaric2010bayesian}, using regularization terms that couple task parameters~\cite{evgeniou2004regularized,Zhong2012a}, and finding mixtures of experts that can be shared across tasks~\cite{Bakker2003}.

Where Bayesian MTL methods aim to find an appropriate bias to share among all task models, transformation methods seek to make one dataset look like another, often in a transfer learning setting. This can be accomplished with distribution matching~\cite{Bickel2008}, inter-task mapping \cite{taylor2007transfer}, or manifold alignment techniques~\cite{wang2009general,ham2005semisupervised,BouAmmar2015Unsupervised}. 

Both the Bayesian strategy of discovering biases and the shared spaces often used in transformation techniques are implicitly connected to methods that learn shared knowledge representations for MTL. For example, the original MTL framework developed by Caruana~\citeyear{Caruana1997} and later variations~\cite{baxter2000model} capture task relationships by sharing hidden nodes in neural networks that are trained on multiple tasks.  
Related work in dictionary learning techniques for MTL \cite{Maurer2013,Kumar2012} factorize the learned models into a shared latent dictionary over the model space to facilitate transfer. Individual task models are then captured as sparse representations over this dictionary; the task relationships are captured in these sparse codes.  
The Efficient Lifelong Learning Algorithm (ELLA) framework \cite{Ruvolo2013} used this same approach of a shared latent dictionary, trained online, to facilitate transfer as tasks arrive consecutively.  The ELLA framework was first created for regression and classification \cite{Ruvolo2013}, and later developed for policy gradient reinforcement learning (PG-ELLA) \cite{Ammar2014a,bouammar2015autonomous}.  
Other approaches that extend MTL to online settings also exist \cite{cavallanti2010linear}. Saha~\etal~\citeyear{saha2011online} use a task interaction matrix to model task relations online and Dekel~\etal~\citeyear{dekel2006online} propose a shared global loss function that can be minimized as tasks arrive. 

However, \emph{all} these methods use task data to characterize the task relationships---this explicitly requires training on the data from each task in order to perform transfer. 
Instead of relying solely on the tasks' training data, several works have explored the use of high-level task descriptors to model the inter-task relationships in MTL and transfer learning settings.  Task descriptors have been used in combination with neural networks \cite{Bakker2003} to define a task-specific prior and to control the gating network between individual task clusters.  Bonilla~\etal~\citeyear{bonilla2007kernel} explore similar techniques for multi-task kernel machines, using task features in combination with the data for a gating network over individual task experts to augment the original task training data.  These papers focus on multi-task classification and regression in batch settings where the system has access to the data and features for all tasks, in contrast to our study of task descriptors for lifelong learning over consecutive tasks. We use coupled dictionary learning to link the task description space with the task's parameter space. This idea was originally used in image processing \cite{yang2010image} and was recently explored in the machine learning literature \cite{xu2016multi}. The core idea is that two feature spaces can be linked through two dictionaries which are coupled by a joint sparse representation.  

In the work most similar to our problem setting, Sinapov~\etal~\citeyear{Sinapov2015} use task descriptors to estimate the transferability between each pair of tasks for transfer learning.  Given the descriptor for a new task, they identify the source task with the highest predicted transferability, and use that source task for a warm start in reinforcement learning (RL).  Though effective, their approach is computationally expensive, since they estimate the transferability for every task pair through repeated simulation.  Their evaluation is also limited to a transfer learning setting, and they do not consider the effects of transfer over consecutive tasks or updates to the transferability model, as we do in the lifelong setting.  
 
Our work is also related to zero-shot learning, which seeks to successfully label out-of-distribution examples, often through means of learning an underlying representation that extends to new tasks and using outside information that appropriately maps to the latent space \cite{Palatucci2009,Socher2013}. The Simple Zero-Shot method by Romera-Paredes and Torr~\citeyear{romera2015embarrassingly} also uses task descriptions. Their method learns a multi-class linear model, and factorizes the linear model parameters, assuming the descriptors are coefficients over a latent basis to reconstruct the models.  Our approach assumes a more flexible relationship: that both the model parameters and task descriptors can be reconstructed from separate latent bases that are coupled together through their coefficients. In comparison to our lifelong learning approach, the Simple Zero-Shot method operates in an offline multi-class setting.

\section{Background}
Our proposed framework for lifelong learning with task descriptors supports both supervised learning (classification and regression) and reinforcement learning settings. For completeness, we briefly review these learning paradigms here.

\subsection{Supervised Learning}
Consider a standard batch supervised learning setting. Let $\bm{x} \in \mathcal{X} \subseteq \Reals^{d}$ be a $d$-dimensional vector representing a single data instance with a corresponding label $y \in \mathcal{Y}$. Given a set of $n$ sample observations $\bm{X} = \{ \bm{x}_1, \bm{x}_2, \dots, \ \bm{x}_n \}$ 
with corresponding labels $\bm{y} = \{y_{1}, y_{2}, \dots, y_{n}\}$,
the goal of supervised learning is to learn a function $f_{\bf\theta} : \mathcal{X} \mapsto \mathcal{Y}$ that labels inputs $\bm{X}$ with their outputs $\bm{y}$ and generalizes well to unseen observations. 

In {\bf regression} tasks, the labels are assumed to be real-valued (i.e., $\mathcal{Y} = \Reals$).  In {\bf classification} tasks,  the labels are a set of discrete classes; for example, in binary classification, $\mathcal{Y} = \{+1, -1\}$. We assume that the learned model for both paradigms $f_{\bm{\theta}}$ can be parameterized by a vector $\bm{\theta}$.  The model is then trained to minimize the average loss over the training data between the model's predictions and the given target labels:
\begin{align*}
	\arg\min_{\bm{\theta}} \frac{1}{n} \sum_{i=1}^{n} \mathcal{L}  \big(  f(\bm{x}_i, \bm{\theta}) , y_{i} \big) + \mathcal{R}(f_{\bm{\theta}}) \enspace ,
\end{align*}
where $\mathcal{L}(\cdot)$ is generally assumed to be a convex metric, and $\mathcal{R}(\cdot)$ regularizes the learned model.  The form of the model $f$, loss function $\mathcal{L}(\cdot)$, and regularization method varies between learning methods.  This formulation encompasses a number of parametric learning methods, including linear regression and logistic regression.

\subsection{Reinforcement Learning}
 
A reinforcement learning (RL) agent selects sequential actions in an environment to maximize its expected return. An RL task is typically formulated as a Markov Decision Process (MDP) $\langle \mathcal{X}, \mathcal{A}, P, R, \gamma  \rangle$, where $\mathcal{X}$ %
is the set of states, and $\mathcal{A}$ is the set of actions \mbox{that the agent may execute,} \mbox{$P: \mathcal{X} \times \mathcal{A} \times \mathcal{X} \rightarrow [0,1]$} is the state transition probability describing the systems dynamics, $R: \mathcal{X} \times \mathcal{A} \times \mathcal{X} \rightarrow \Reals $ is the reward function, and $\gamma \in [0,1)$ is the discount assigned to rewards over time. At time step $h$, the agent is in state $\bm{x}_h\in\mathcal{X}$ and chooses an action $\bm{a} \in \mathcal{A}$ according to policy \mbox{$\pi: \mathcal{X} \times \mathcal{A} \mapsto [0,1]$}, which is represented as a function defined by a vector of control parameters \mbox{$\bm{\theta} \in \Reals^d$}. The agents then receives reward $r_h$ according to $R$ and transitions to state $\bm{x}_{h+1}$ according to $P$. This sequence of states, actions, and rewards is given as a trajectory \mbox{$\bm{\tau} = \{(\bm{x}_1, \bm{a}_1, r_1), \ldots, (\bm{x}_H, \bm{a}_H, r_H)\}$} over a horizon $H$. The goal of RL is to find the optimal policy $\pi^*$ with parameters $\bm{\theta}^*$ that maximizes the expected reward.  However, learning an individual task still requires numerous trajectories, motivating the use of transfer to reduce the number of interactions with the environment. 

Policy Gradient (PG) methods \cite{sutton1999policy}, which we employ as our base learner for RL tasks, are a class of RL algorithms that are effective for solving high dimensional problems with continuous state and action spaces, such as robotic control \cite{peters2008natural}. %
The goal of PG is to optimize the expected average return: 
\mbox{$%
 	\mathcal{J}(\bm{\theta}) = E \left[ \frac{1}{H} \sum_{h=1}^H r_h \right] = \int_{\mathbb{T}} p_{\bm{\theta}}(\bm{\tau})\mathfrak{R}(\bm{\tau})d\bm{\tau}
 $}, %
where $\mathbb{T}$ is the set of all possible trajectories, the average reward on trajectory $\bm{\tau}$ is given by $\mathfrak{R}(\bm{\tau}) = \frac{1}{H}\sum_{h=1}^{H} r_h$, and $p_{\bm{\theta}}(\bm{\tau}) = P_0(\bm{x}_1) \prod_{h=1}^{H} p(\bm{x}_{h+1} \mid \bm{x}_{h},\bm{a}_{h}) \ \pi(\bm{a}_h \mid \bm{x}_{h})$ is the probability of $\bm{\tau}$ under an initial state distribution \mbox{$P_0:\mathcal{X} \mapsto [0,1]$}. Most PG methods (e.g., episodic REINFORCE~\cite{williams1992simple}, PoWER~\cite{kober2009policy}, and Natural Actor Critic~\cite{peters2008natural}) optimize the policy by employing supervised function approximators to maximize a lower bound on the expected return of  $\mathcal{J}(\bm{\theta})$.  %
This optimization is carried out by generating trajectories using the current policy $\pi_{\bm{\theta}}$, and then comparing the result with a new policy $\pi_{\tilde{\bm{\theta}}}$. Jensen's inequality can then be used to lower bound the expected return \cite{kober2009policy}:  
\begin{align*}
\log\mathcal{J}\!\left(\thetaprime\right) &= \log \int_{\trajectories} p_{\thetaprime}(\taut) \ \mathfrak{R}(\taut) \diff\taut\\
&=\log \int_{\trajectories} \frac{p_{\bm{\theta}}(\taut)}{p_{\bm{\theta}}(\taut)} \ p_{\thetaprime}(\taut) \ \mathfrak{R}(\taut) \diff\taut\\
&\geq \int_{\trajectories} p_{\bm{\theta}}(\taut) \ \mathfrak{R}(\taut) \log\frac{p_{\thetaprime}(\taut)}{p_{\bm{\theta}}(\taut)} \diff\taut+\text{constant} \\
&\propto -\mathfrak{D}_{\text{KL}}\bigl(p_{\bm{\theta}}(\taut)\ \mathfrak{R}(\taut) \mid\mid p_{\thetaprime}(\taut)\!\bigr)=\objJtheta\!\left(\thetaprime\right) \enspace , 
\end{align*}
where $\ \mathfrak{D}_{\text{KL}}\bigl(p(\taut) \mid\mid q(\taut)\bigr)= \displaystyle{\int_\trajectories} p(\taut)\log\frac{p(\taut)}{q(\taut)} \diff\taut$. 
This is equivalent to minimizing the KL divergence between the reward-weighted trajectory distribution of $\pi_{\bm{\theta}}$ and the trajectory distribution $p_{\thetaprime}$ of the new policy $\pi_{\thetaprime}$. 

In our work, we treat the term $\objJtheta\!\left(\thetaprime\right)$ similar to the loss function $\mathcal{L}$ of a classification or  regression task. Consequently, both supervised learning tasks and RL tasks can be modeled in a unified framework, where the goal is to minimize a convex loss function. %

\subsection{Lifelong Machine Learning}

\begin{figure}[t!]
    \centering
    \includegraphics[width=0.9\textwidth,clip,trim=1.35in 0.6in 1.35in 1.5in]{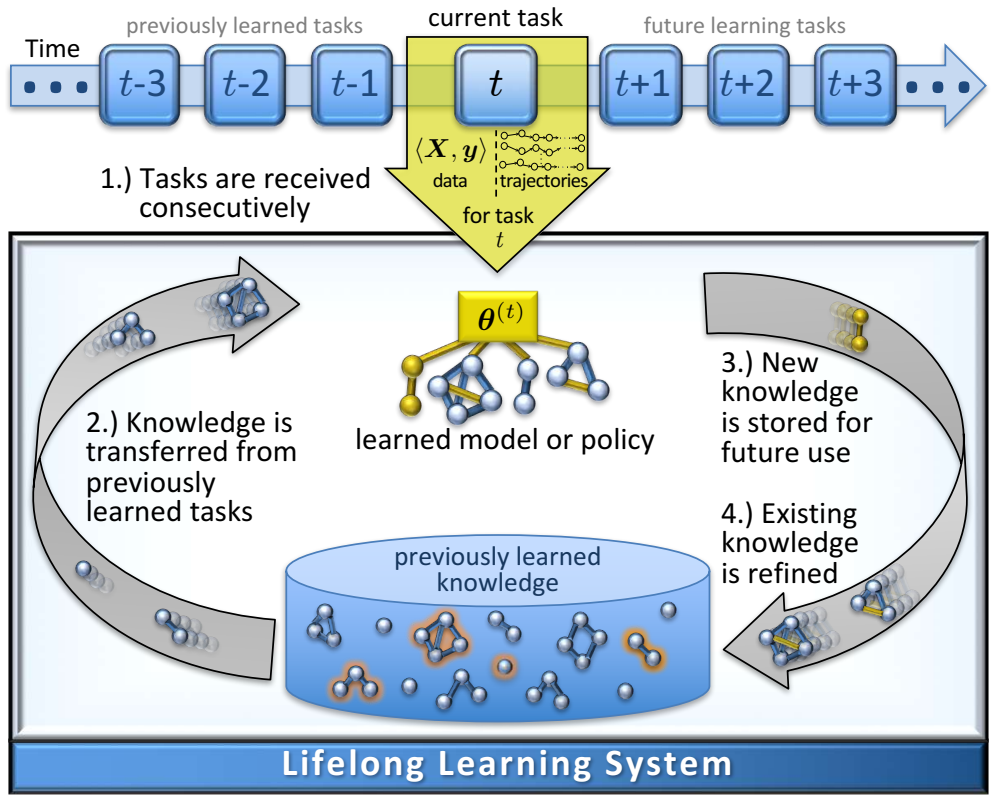}
         \caption{The lifelong machine learning process. As a new task arrives, knowledge accumulated from previous tasks is selectively transferred to the new task to improve learning. Newly learned knowledge is then stored for future use.}
         \label{fig:LL}
\end{figure}

In a lifelong learning setting \cite{Thrun1996,Ruvolo2013}, a learner faces multiple, consecutive tasks and must rapidly learn each new task by building upon its previous experience.  The learner may encounter a previous task at any time, and so must optimize performance across all tasks seen so far.  A priori, the agent does not know the total number of tasks $T_{\mathrm{max}}$, the task distribution, or the task order.

At time $t$, the lifelong learner encounters task $\Task{t}$.  In this paper, all tasks are either regression problems $\Task{t} = \langle \Xt{t}, \yt{t} \rangle$, classification problems $\Task{t} = \langle \Xt{t}, \yt{t} \rangle$ or reinforcement learning problems specified by an MDP $\langle \mathcal{X}^{(t)}, \mathcal{A}^{(t)}, P^{(t)}, R^{(t)}, \gamma^{(t)}  \rangle$. Note that we do not mix the learning paradigms---a lifelong learning agent will only face one type of learning task during its lifetime.  The agent will learn each task consecutively, acquiring training data (i.e., trajectories or samples) in each task before advancing to the next.  
The agent's goal is to learn the optimal models $\{f^*_{\Th{1}}, \ldots, f^*_{\Th{T}}\}$ or policies $\{\pi^*_{\Th{1}}, \ldots, \pi^*_{\Th{T}}\}$ with corresponding parameters $\{\Th{1}, \ldots, \Th{T}\}$, where $T$ is the number of unique tasks seen so far ($1 \leq T \leq T_{\mathrm{max}}$).  Ideally, knowledge learned from previous tasks $\{\Task{1}, \dots, \Task{T-1}\}$ should accelerate training and improve performance on each new task $\Task{T}$. Also, the lifelong learner should scale effectively to large numbers of tasks, learning each new task rapidly from minimal data. The lifelong learning framework is depicted in Figure \ref{fig:LL}.

\begin{figure}[t]
    \centering
    \includegraphics[height=2in,clip,trim=4in 3in 4in 2.9in]{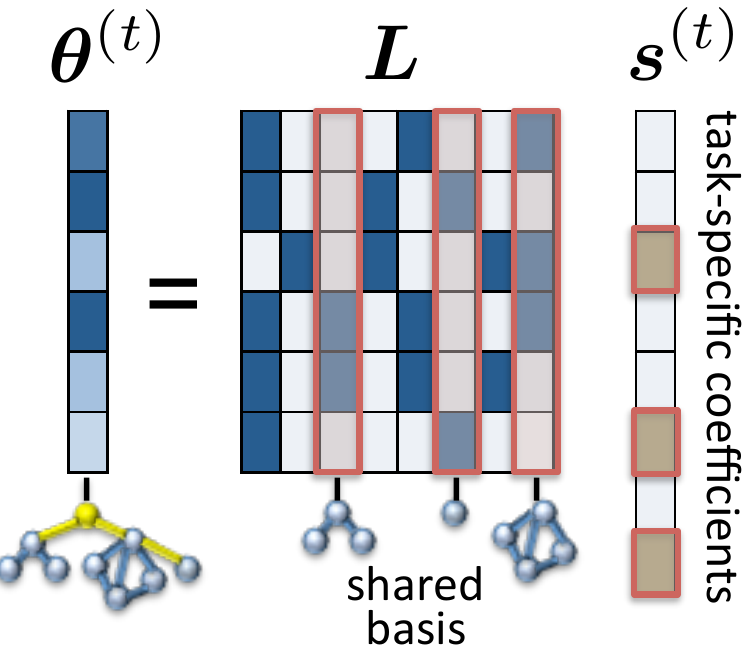}
    \caption{The task specific model (or policy) parameters $\Th{t}$ are factored into a shared knowledge repository $\bm{L}$ and a sparse code $\st{t}$. The repository $\bm{L}$ stores chunks of knowledge that are useful for multiple tasks, and the sparse code $\st{t}$ extracts the relevant pieces of knowledge for a particular task's model (or policy).}
    \label{fig:ThetaLs}
\end{figure}

The Efficient Lifelong Learning Algorithm (ELLA)~\cite{Ruvolo2013} and PG-ELLA~\cite{Ammar2014a} were developed to operate in this lifelong learning setting for classification/regression and RL tasks, respectively. Both approaches assume the parameters for each task model can be factorized using a shared knowledge base $\bm{L}$, facilitating transfer between tasks.  Specifically, the model parameters for task $\Task{t}$ are given by $\Th{t} = \bm{L}\st{t}$, where $\bm{L}\in \mathbb{R}^{d\times k}$ is the shared basis over the model space, and $\st{t}\in \mathbb{R}^k$ are the sparse coefficients over the basis. This factorization, depicted in Figure \ref{fig:ThetaLs}, has been effective for transfer in both lifelong and multi-task learning~\cite{Kumar2012,Maurer2013}.

Under this assumption, the MTL objective is: 
\begin{equation} \label{eqn:MTLObjective}
	\min_{\bm{L},\bm{S}} \frac{1}{T} \sum_{t=1}^{T} \left[\Loss(\Th{t}) + \mu \| \st{t} \|_1 \right] + \lambda\|\bm{L}\|_\textsf{F}^2 \enspace,
\end{equation}
where $\bm{S} = [\st{1} \cdots\ \st{T}]$ is the matrix of sparse vectors, $\Loss$  is the task-specific loss 
for task $\Task{t}$, and $\|\cdot\|_{\mathsf{F}}$ is the Frobenius norm. The $L_1$ norm is used to approximate the true vector sparsity of $\st{t}$, and $\mu$ and $\lambda$ are regularization parameters. Note that for a convex loss function $\Loss(\cdot)$, this problem is convex in each of the variables $\bm{L}$ and $\bm{S}$. Thus, one can use an alternating optimization approach to solve it in a batch learning setting.
To solve this objective in a lifelong learning setting, Ruvolo and Eaton \citeyear{Ruvolo2013} take a second-order Taylor expansion to approximate the objective around an estimate $\bm{\alpha}^{(t)} \in \Reals^d$ of the single-task model parameters for each task $\Task{t}$, and update only the coefficients $\st{t}$ for the current task at each time step. This process reduces the MTL objective to the problem of sparse coding the single-task policies in the shared basis $\bm{L}$, and enables $\bm{S}$ and $\bm{L}$ to be solved efficiently by the following alternating online update rules that constitute ELLA \cite{Ruvolo2013}:

\begin{align} 
\st{t} &\leftarrow \arg\min_{\bm{s}} \| \bm{\alpha}^{(t)} -\bm{L} \bm{s}\|^2_{\bm{\Gamma}^{(t)}} + \mu \| \bm{s}\|_1 \label{eqn:updaterulest} \\
\bm{A} &\leftarrow \bm{A} + (\bm{s}^{(t)}\bm{s}^{(t)\top})\otimes\bm{\Gamma}^{(t)}  \label{eqn:updateruleA} \\
\bm{b} &\leftarrow \bm{b}  + \text{vec}\left(\bm{s}^{(t)\top}\otimes \left(\bm{\alpha}^{(t) \top} \bm{\Gamma}^{(t)} \right)\right)  \label{eqn:updateruleB}\\
\bm{L} &\leftarrow \text{mat} \biggl( \biggl( \frac{1}{T}\bm{A} +\lambda\bm{I}_{kd}\biggr)^{-1}\frac{1}{T}\bm{b} \biggr)  \label{eqn:updateruleL} \enspace,
\end{align} 
where $\|\bm{v}\|^2_{\bm{A}} = \bm{v}^{\!\top}\!\!\bm{A}\bm{v}$, the symbol $\otimes$ denotes the Kronecker product, $\bm{\Gamma}^{(t)}$ is the Hessian of the loss $\Loss(\bm{\alpha}^{(t)})$, $\bm{I}_{m}$ is the $m \times m$ identity matrix, %
$\bm{A}$ is initialized to a $kd \times kd$ zero matrix, and $\bm{b} \in \Reals^{kd}$ is initialized to zeros.

This was extended to handle reinforcement learning by Bou Ammar~\etal~\citeyear{Ammar2014a} via  approximating the RL multi-task objective by first substituting in the convex lower-bound to the PG objective  $\mathcal{J}(\bm{\alpha}^{(t)})$ in order to make the optimization convex.

While these methods are effective for lifelong learning, this approach requires training data to estimate the model for each new task before the learner can solve it. Our key idea is to eliminate this restriction by incorporating task descriptors into lifelong learning, enabling zero-shot transfer to new tasks. That is, upon learning a few tasks, future task models can be predicted solely using task descriptors. %

\section{Lifelong Learning with Task Descriptors} \label{sect:Approach}
\begin{figure}[t!]
    \centering
     \includegraphics[width=0.9\textwidth,clip,trim=1.35in 0.6in 1.35in 1.5in]{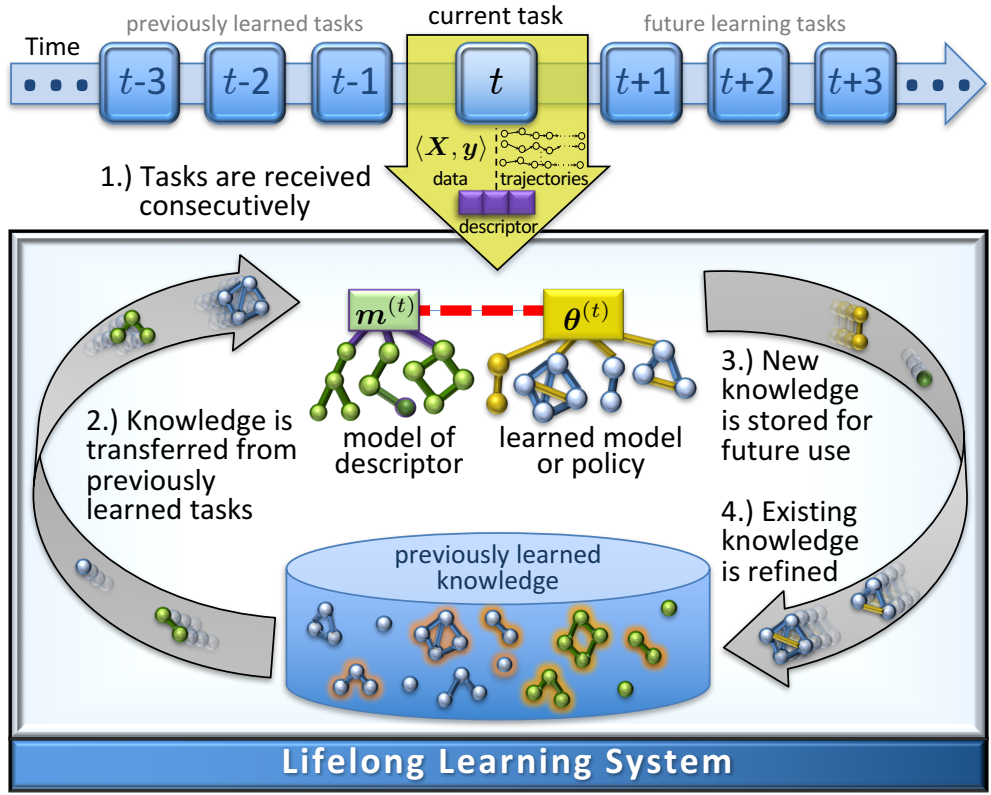}
    \label{fig:TaDeLL}
         \caption{The lifelong machine learning process with task descriptions. A model of task descriptors is added into the lifelong learning framework and couple with the learned model. Because of the learned coupling between model and description, the model for a new task can be predicted from the task description.}
\end{figure}
\subsection{Task Descriptors} 

While most MTL and lifelong learning methods use task training data to model inter-task relationships, high-level descriptions can describe task differences. For example, in multi-task medical domains, patients are often grouped into tasks by demographic data and disease presentation \cite{oyen2012leveraging}. %
In control problems, the dynamical system parameters (e.g., the spring, mass, and damper constants in a spring-mass-damper system) describe the task. Descriptors can also be derived from external sources, such as text descriptions ~\cite{pennington2014glove,Huang2012} or Wikipedia text associated with the task ~\cite{Socher2013}.  

To incorporate task descriptors into the learning procedure, we assume that each task $\Task{t}$ has an associated descriptor $\mt{t}$ that is given to the learner upon first presentation of the task.  The learner has no knowledge of future tasks, or the distribution of task descriptors.  The descriptor is represented by a feature vector $\Mt{t} \in \Reals^{d_m}$, where $\phi(\cdot)$ performs feature extraction and (possibly) a non-linear basis transformation on the features.  We make no assumptions on the uniqueness of $\Mt{t}$, although in general tasks will have different descriptors.\footnote{This raises the question of what descriptive features to use, and how task performance will change if some descriptive features are unknown.  We explore these issues in Section~\ref{whichfeatures}.}  In addition, each task also has associated training data $\bm{X}^{(t)}$ to learn the model; in the case of RL tasks, the data consists of trajectories that are dynamically acquired by the agent through experience in the environment.

We incorporate task descriptors into lifelong learning via sparse coding with a coupled dictionary, enabling the descriptors and learned models to augment each other. %
In an earlier version of this work, we focused on RL tasks \cite{isele2016task}; here, we more fully explore the range %
of our approach, showing how it can be applied to regression, classification, and RL problems.

\subsection{Coupled Dictionary Optimization}

As described previously, many multi-task and lifelong learning approaches have found success with factorizing the policy parameters $\Th{t}$ for each task as a sparse linear combination over a shared basis:  $\Th{t} = \bm{L}\st{t}$.  In effect, each column of the shared basis $\bm{L}$ serves as a reusable model or policy component representing a cohesive chunk of knowledge.  In lifelong learning, the basis $\bm{L}$ is refined over time as the system learns more tasks.  The coefficient vectors $\bm{S} = [\st{1} \ldots \st{T}]$ encode the task policies in this shared basis, providing an embedding of the tasks based on how their policies share knowledge.

\begin{figure}[t!]
    \centering
    \includegraphics[height=2.5in,clip,trim=1.4in 2.4in 1.4in 2.4in]{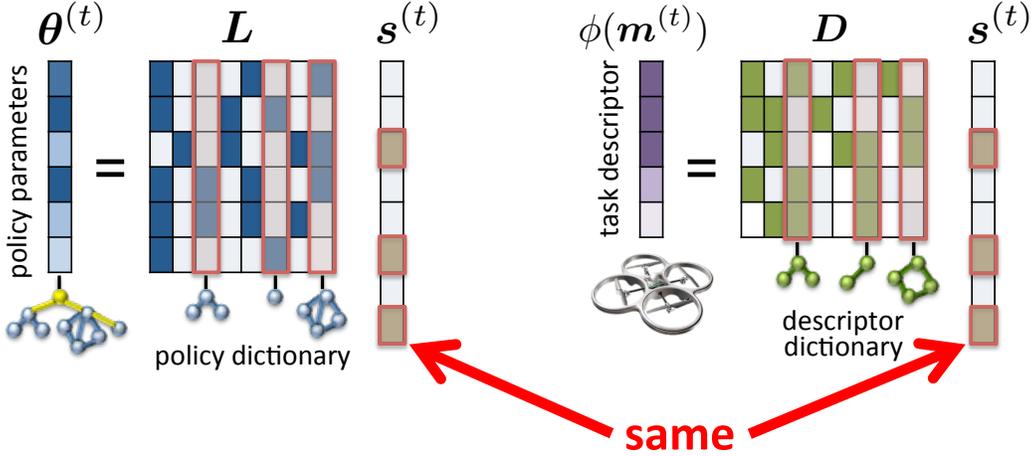}
    \caption{The coupled dictionaries of TaDeLL, illustrated on an RL task. Policy parameters $\Th{t}$ are factored into $\bm{L}$ and $\st{t}$ while the task description $\Mt{t}$ is factored into $\bm{D}$ and $\st{t}$. Because we force both dictionaries to use the same sparse code $\st{t}$, the relevant pieces of information for a task become coupled with the description of the task.}
    \label{fig:LsDs}
\end{figure}

We make a similar assumption about the task descriptors---that the descriptor features $\Mt{t}$ can be linearly factorized\footnote{This is potentially non-linear w.r.t~$\mt{t}$, since $\phi$ can be non-linear.} using a latent basis $\bm{D} \in \Reals^{d_m \times k}$ over the descriptor space.  This basis captures relationships among the descriptors, with coefficients that similarly embed tasks based on commonalities in their descriptions.  From a co-view perspective \cite{yu2014discriminative}, both the policies and descriptors provide information about the task, and so each can augment the learning of the other.  Each underlying task is common to both views, and so we seek to find task embeddings that are consistent for {\em both} the policies and their corresponding task descriptors.  As depicted in Figure \ref{fig:LsDs}, we can enforce this by coupling the two bases $\bm{L}$ and $\bm{D}$, sharing the same coefficient vectors $\bm{S}$ to reconstruct both the policies and descriptors. Therefore, for task $\Task{t}$,
\begin{align} \label{eqn:FactorizationModel}
\Th{t} &= \bm{L}\st{t} & \Mt{t} &= \bm{D}\st{t} \enspace . 
\end{align}

To optimize the coupled bases $\bm{L}$ and $\bm{D}$ during the lifelong learning process, we employ techniques for coupled dictionary optimization from the sparse coding literature~\cite{yang2010image}, which optimizes the dictionaries for multiple feature spaces that share a joint sparse representation.  %
This notion of coupled dictionary learning has led to high performance algorithms for image super-resolution \cite{yang2010image}, allowing the reconstruction of high-res images from low-res samples, and for multi-modal retrieval \cite{zhuang2013supervised} %
and cross-domain retrieval \cite{yu2014discriminative}. 
The core idea is that features in two independent subspaces can have the same representation in a third subspace.

Given the factorization in Eq.~\ref{eqn:FactorizationModel}, we can re-formulate the multi-task objective (Eq.~\ref{eqn:MTLObjective}) for the coupled dictionaries as  
\begin{equation}\label{eqn:MTLObjectiveCoupledDictionaries}
\begin{split}
	\min_{\bm{L},\bm{D},\bm{S}} \ \frac{1}{T} \sum_t \biggl[&\!\Loss\left(\Th{t} \right) + \rho\left\|\Mt{t} - \bm{D}\st{t} \right\|_2^2   + \mu \left\| \st{t} \right\|_1 \biggr] + \lambda(\|\bm{L}\|_\textsf{F}^2  + \|\bm{D}\|_\textsf{F}^2) \enspace ,
\end{split}
\end{equation}
where $\rho$ balances the model's or policy's fit to the task descriptor's fit.

To solve Eq.~\ref{eqn:MTLObjectiveCoupledDictionaries} online, we approximate $\Loss(\cdot)$ by a second-order Taylor expansion around $\At{t}$, the minimizer for the 
single-task learner.
In reinforcement learning, $\pi_{\At{t}}$ is the single-task policy for $\Task{t}$ based on the observed trajectories \cite{Ammar2014a}. In supervised learning, $\At{t}$ is the single-task model parameters for $\Task{t}$ \cite{Ruvolo2013}. This step leads to a unified simplified formalism that is independent of the learning paradigm (i.e., classification, regression, or RL). Approximating  Eq.~\ref{eqn:MTLObjectiveCoupledDictionaries} leads to 
\begin{align}
	\min_{\bm{L},\bm{D},\bm{S}}  \frac{1}{T}\! \sum_t \biggl[\!\left\|\At{t} \!-\! \bm{L}\st{t} \right\|_{\bm{\Gamma}^{(t)}}^2 \!+ \rho\left\|\Mt{t} \!-\! \bm{D}\st{t} \right\|_2^2    + \mu \left\| \st{t} \right\|_1 \biggr] \!+ \lambda(\|\bm{L}\|_\textsf{F}^2  + \|\bm{D}\|_\textsf{F}^2) \enspace .
\label{eqn:MTLObjectiveCoupledDictionariesSparseCoded}
\end{align}
We can merge pairs of terms in Eq.~\ref{eqn:MTLObjectiveCoupledDictionariesSparseCoded}  by choosing:
\begin{align*}
\ComboVec{t} &\!=\! \begin{bmatrix} \At{t} \\ \Mt{t} \end{bmatrix} & 
\ComboDict &\!=\! \begin{bmatrix}\bm{L} \\ \bm{D} \end{bmatrix} &
\bm{A}^{(t)} &\!=\! \begin{bmatrix} \bm{\Gamma}^{(t)} & \bm{0} \\ \bm{0} & \rho\bm{I}_{d_m} \end{bmatrix}
\enspace ,
\end{align*}
where $\bm{0}$ is the zero matrix, letting us rewrite \eqref{eqn:MTLObjectiveCoupledDictionariesSparseCoded} concisely as
\begin{align}
\!\!\min_{\ComboDict,\bm{S}}  \frac{1}{T} \! \sum_t \biggl[\!\left\|\ComboVec{t} \!-\! \ComboDict \st{t} \right\|_{\bm{A}^{(t)}}^2 \!\!+ \mu\! \left\| \st{t} \right\|_1 \!\biggr] \!+\! \lambda\|\ComboDict\|_\textsf{F}^2 \enspace .\!
\label{eqn:MTLObjectiveCoupledDictionariesSparseCodedConcise}
\end{align}
This objective can now be solved efficiently online, as a series of per-task update rules given in Algorithm~\ref{algo:TaDeLL}, which we call TaDeLL (Task Descriptors for Lifelong Learning). $\bm{L}$ and $\bm{D}$ are updated independently using Equations \ref{eqn:updateruleA}--\ref{eqn:updateruleL}, following a recursive construction based on an eigenvalue decomposition.

\begin{algorithm}[tb!]
\caption{\ TaDeLL (k, $\lambda$, $\mu$)}
\label{algo:TaDeLL}
\begin{algorithmic}[1]
\State $\bm{L} \gets  \bm{\text{RandomMatrix}}_{d,k}, \ \bm{D} \gets  \bm{\text{RandomMatrix}}_{m,k}$
\While {some task $\left(\Task{t}, \Mt{t}\right)$ is available}

\State $\mathbb{T}^{(t)} \gets$ collectData($\Task{t}$)

\State Compute $\At{t}$ and $\bm{\Gamma}^{(t)}$ from $\mathbb{T}^{(t)}$
\State $\st{t} \gets \arg\min_{\bm{s}} \left\|\ComboVec{t} \!-\! \ComboDict \bm{s} \right\|_{\bm{A}^{(t)}}^2 + \mu\! \left\| \bm{s} \right\|_1$
\State $\bm{L} \gets$ update$L(\bm{L}, \st{t}, \At{t}, \bm{\Gamma}^{(t)}, \lambda)$ \hspace{43pt} Eq. \ref{eqn:updateruleA}--\ref{eqn:updateruleL}
\State $\bm{D} \gets$ update$D(\bm{D},\st{t}, \Mt{t}, \rho \bm{I}_{d_m}, \lambda)$ \hspace{12pt} Eq. \ref{eqn:updateruleA}--\ref{eqn:updateruleL}
\State \textbf{for} $t \in \{1, \ldots, T\}$ \textbf{do}:\ \ $\Th{t} \gets \bm{L} \st{t}$ 
\EndWhile
\end{algorithmic}
\end{algorithm}

For the sake of clarity, we now explicitly state the differences between using TaDeLL for RL problems and for classification and regression problems. 
In an RL setting, at each timestep TaDeLL receives a new RL task and samples trajectories for the new task. We use the single-task policy as computed using a twice-differentiable policy gradient method as $\bm{\alpha}^{(t)}$. The Hessian $\Gamma^{(t)}$, calculated around the point $\bm{\alpha}^{(t)}$, is derived according to the particular policy gradient method being used. Bou Ammar \etal \citeyear{Ammar2014a} derive it for the cases of Episodic REINFORCE and Natural Actor Critic. The reconstructed $\bm{\theta}^{(t)}$ is then used as the policy for the task $\Task{t}$.

In the case of classification and regression, at each time step TaDeLL observes a labeled training set $(\bm{X}^{(t)}, \bm{y}^{(t)})$ for task $\Task{t}$, where $\bm{X}^{(t)} \subseteq \Reals^{n_t \times d}$.  For classification tasks, $\bm{y}^{(t)} \in \{+1, -1\}^{n_t}$, and for regression tasks, $\bm{y}^{(t)} \in \Reals^{n_t}$. We then set $\bm{\alpha}^{(t)}$ to be the parameters of a single-task model trained via classification or regression (e.g., logistic or linear regression) on that data set. $\bm{\Gamma}^{(t)}$ is set to be the Hessian of the corresponding loss function around the single-task solution $\bm{\alpha}^{(t)}$, and the reconstructed $\bm{\theta}^{(t)}$ is used as the model parameters for the corresponding classification or regression problem.

\subsection{Zero-Shot Transfer Learning}

In a lifelong setting, when faced with a new task, the agent's goal is to learn an effective policy for that task as quickly as possible.  At this stage, previous multi-task and lifelong learners incurred a delay before they could produce a decent policy, since they needed to acquire data from the new task in order to identify related knowledge and train the new policy via transfer.  

Incorporating task descriptors enables our approach to predict a policy for the new task immediately, given {\em only} the descriptor.  This ability to perform zero-shot transfer is enabled by the use of coupled dictionary learning, which allows us to observe a data instance in one feature space (i.e., the task descriptor), and then recover its underlying latent signal in the other feature space (i.e., the policy parameters) using the dictionaries and sparse coding. %

Given only the descriptor $\mt{t_{\mathit{new}}}$ for a new task $\Task{t_{\mathit{new}}}$, we can estimate the embedding of the task in the latent descriptor space via LASSO on the learned dictionary $\bm{D}$:
\begin{align}
\tilde{\bm{s}}^{(t_{\mathit{new}})} \gets \arg\min_{\bm{s}} \left\|\Mt{t} - \bm{D} \bm{s} \right\|_2^2 + \mu \left\| \bm{s} \right\|_1 \enspace.
\end{align}
Since the estimate given by $\tilde{\bm{s}}^{(t_{\mathit{new}})}$ also serves as the coefficients over the latent policy space $\bm{L}$, we can immediately predict a policy for the new task as: $\tilde{\bm{\theta}}^{(t_{\mathit{new}})} = \bm{L} \tilde{\bm{s}}^{(t_{\mathit{new}})}$.  This zero-shot transfer learning procedure is given as Algorithm~\ref{alg:ZeroShot}.  

\begin{algorithm}[t!]
\caption{\ Zero-Shot Transfer to a New Task $\Task{t_{\mathit{new}}}$ }
\label{alg:ZeroShot}
\begin{algorithmic}[1]
\State \textbf{Inputs:} task descriptor $\mt{t_{\mathit{new}}}$, learned bases $\bm{L}$ and $\bm{D}$
\State $\tilde{\bm{s}}^{(t_{\mathit{new}})} \gets \arg\min_{\bm{s}} \left\|\Mt{t_{\mathit{new}}} - \bm{D} \bm{s} \right\|_2^2 + \mu \left\| \bm{s} \right\|_1$
\State $\tilde{\bm{\theta}}^{(t_{\mathit{new}})} \gets \bm{L} \tilde{\bm{s}}^{(t_{\mathit{new}})}$
\State \textbf{Return:}  $\pi_{\tilde{\bm{\theta}}^{(t_{\mathit{new}})}}$
\end{algorithmic}
\end{algorithm}

\section{Theoretical Analysis}
\label{sect:theory}

This section examines theoretical issues related to incorporating task descriptors into lifelong learning via coupled dictionaries. We start by outlining theory to support why the inclusion of task features can improve performance of the learned policies and enable zero-shot transfer to new tasks safely. We then prove the convergence of TaDeLL.   
A full sample complexity analysis is beyond the scope of this paper, and, indeed, remains an open problem for zero-shot learning \cite{romera2015embarrassingly}.

\subsection{Connections to Mutual Coherence in Sparse Coding}

To analyze the policy improvement, since the policy parameters are factored as  $\Th{t}=\bm{L}\st{t}$, we proceed by showing that incorporating the descriptors through coupled dictionaries can improve both $\bm{L}$ and $\bm{S}$. Note that learning these dictionaries faster means faster knowledge transfer and more accurate task prediction.  In this analysis, we employ the 
concept of {\em mutual coherence}, which has been studied extensively in the sparse recovery literature. \cite{donoho2006stable}
Mutual coherence measures the similarity of a dictionary's elements as
\mbox{$M(\bm{Q})=\max_{1\le i\neq j\le k}\left(\frac{|\bm{q}_i^\top\bm{q}_j|}{\|\bm{q}_i\|_2\|\bm{q}_j\|_2}\right) \in [0, 1]$}, 
where $\bm{q}_i$ is the $i^{th}$ column of a dictionary $\bm{Q}\in\mathbb{R}^{d\times k}$. If $M(\bm{Q})=0$, then $\bm{Q}$ is an invertible orthogonal matrix and so sparse recovery can be solved directly by inversion; $M(\bm{Q})=1$ implies that $\bm{Q}$ is not full rank and a poor dictionary. 
Intuitively, low mutual coherence indicates that the dictionary's columns  are considerably different, and thus such a ``good'' dictionary can represent wider range of  tasks, potentially yielding more knowledge transfer.  This intuition is shown in the following:
\begin{theorem}
\cite{donoho2006stable}
\label{thrm:coherence}
Suppose we have noisy observations $\hat{\bm{\theta}}$  of the linear system  $\bm{\theta}=\bm{Q}\bm{s}$, such that $\|\hat{\bm{\theta}} - \bm{\theta}\|_2 \leq \epsilon$. Let $\bm{s}^*$ be a solution to the system, and let $K = \|\bm{s}\|_0$. If \mbox{$K<0.5(1+M(\bm{Q})^{-1})$}, then $\bm{s}^*$ is the unique sparsest solution of the system. Moreover, if $\bm{s}^+$ is the LASSO solution
for the system from the noisy observations, then:
$\|\bm{s}^*-\bm{s}^+\|_2^2 \leq \frac{4\epsilon^2}{1-M(\bm{Q})(4K-1)}$.
\end{theorem}
\noindent Therefore, an $\bm{L}$ with low mutual coherence would lead to more stable solutions of the $\Th{t}$'s against inaccurate single-task estimates of the policies (the $\bm{\alpha}^{(t)}$'s).  We next show that our approach likely lowers the mutual coherence of $\bm{L}$.

TaDeLL alters the problem from training $\bm{L}$ to training the coupled dictionaries $\bm{L}$ and $\bm{D}$ (contained in $\ComboDict$).  Let $\bm{s}^{*(t)}$ be the solution to Eq.~\ref{eqn:MTLObjective} for task $\Task{t}$, which is unique under sparse recovery theory, so $\left\|\bm{s}^{*(t)}\right\|_0$ remains unchanged for all tasks.  Theorem~\ref{thrm:coherence} implies that, if $M(\ComboDict) < M(\bm{L})$, coupled dictionary learning can
help with a more accurate recovery of the $\st{t}$'s.  To show this, we note that  %
Eq.~\ref{eqn:MTLObjectiveCoupledDictionaries} can also be
derived as a result of an MAP estimate from a Bayesian perspective, enforcing a Laplacian distribution on the $\st{t}$'s and assuming $\bm{L}$ to be a Gaussian matrix with elements drawn i.i.d: $l_{ij} \sim  \mathcal{N}(\bm{0}, \sigma^2)$. 
When considering such a random matrix $\bm{Q}\in\mathbb{R}^{d\times k}$, 
Donoho \& Huo~\citeyear{donoho2001uncertainty} proved that asymptotically $M(\bm{Q})\propto \sqrt{\frac{\log(dk)}{d}}$ as $d \rightarrow
\infty$.  Using this as an estimate for $M(\bm{L})$ and $M(\ComboDict)$, since %
incorporating task
descriptors increases $d$, asymptotically $M(\ComboDict) < M(\bm{L})$, implying that TaDeLL learns a superior dictionary. Moreover, if $M(\bm{D}) \le M(\bm{L})$,
the theorem implies we can use $\bm{D}$ alone to recover the task policies through zero-shot transfer.

To show that task features can also improve the sparse recovery, we rely on the following theorem about LASSO:
\begin{theorem}
\cite{negahban2009unified}
\label{thrm:lasso}
Let $\bm{s}^*$ be a unique solution to the system $\bm{\theta}=\bm{Q}\bm{s}$ with $\|\bm{s}\|_0=K$ and $\bm{Q} \in \mathbb{R}^{d\times k}$. If $\bm{s}^+$ is the LASSO solution
for the system from noisy observations, then with high probability: \mbox{$\|\bm{s}^*-\bm{s}^+\|_2 \leq c'\sqrt{K\frac{\log k}{d}}$}  \enspace, 
where the constant $c'\in\mathbb{R}^+$ depends on properties of the linear system and observations. 
\end{theorem} 
This theorem shows that the error reconstruction for LASSO is proportional to $\frac{1}{\sqrt{d}}$. When we incorporate the descriptor through $\ComboVec{t}$, the RHS denominator increases from $d$ to $(d+d_m)$ while $K$ and $k$ remain constant, yielding a tighter fit.  %
Therefore, task descriptors can improve learned dictionary quality and sparse recovery accuracy.   %
To ensure an equivalently tight fit for $\st{t}$ %
using either policies or descriptors, Theorem~\ref{thrm:lasso} suggests it should be that $d_m \geq d$ to ensure that zero-shot learning yields similarly tight estimates of $\st{t}$.

\subsection{Theoretical Convergence of TaDeLL}  \label{sect:convergence}

In this section, we prove the convergence of TaDeLL, showing that the learned dictionaries become increasingly stable as it learns more tasks.  We build upon the theoretical results from Bou Ammar~\etal~\citeyear{Ammar2014a} and Ruvolo \& Eaton~\citeyear{Ruvolo2013}, demonstrating that these results apply to coupled dictionary learning with task descriptors, and use them to prove convergence.

Let $\hat{g}_T(\mbox{L})$ represent the sparse coded approximation to the MTL objective, which can be defined as:
\begin{equation*} \label{eqn:ELLAObjective}
 \hat{g}_T(\bm{L})=\frac{1}{T}\sum_{t=1}^T\|\bm{\alpha}^{(t)}\!-\!\bm{L}\st{t}\|^2_{\bm{\Gamma^{(t)}}}+\mu\|\st{t}\|_1+\lambda\|\bm{L}\|_{\mathsf{F}}^2 \enspace 	.
\end{equation*}
 This equation can be viewed as the cost for $\bm{L}$ when the sparse coefficients are kept constant. Let $\bm{L}_T$ be the version of the dictionary $\bm{L}$ obtained after observing $T$ tasks.  Given these definitions, we consider the following theorem:
 \begin{theorem}
 \cite{Ruvolo2013}
 \label{thrm:convergence}
\begin{enumerate}
\item The trained dictionary $\bm{L}$ is stabilized over learning with rate: $\bm{L}_T-\bm{L}_{T-1}=O(\frac{1}{T})$
\item $\hat{g}_T(\bm{L}_T)$ converges almost surely.
\item $\hat{g}_T(\bm{L}_T)-\hat{g}_T(\bm{L}_{T-1})$  converges almost surely to zero.
\end{enumerate}
 \end{theorem}

\noindent This theorem requires two conditions:
\begin{enumerate}
\item The tuples $\bm{\Gamma}^{(t)}$, $\At{t}$ are drawn i.i.d from a distribution with compact support to bound the norms of $\bm{L}$ and $\st{t}$.
\item For all $t$, let $\bm{L}_{\kappa}$ be the subset of the dictionary $\bm{L}_t$, where only columns  corresponding to non-zero element of $\st{t}$ are included.  Then, all eigenvalues of the matrix $\bm{L}_{\kappa}^\transpose\bm{\Gamma}^{(t)}\bm{L}_{\kappa}$ need to be strictly positive.
\end{enumerate}
 
Bou Ammar~\etal~\citeyear{Ammar2014a} show that both of these conditions are met for the lifelong learning framework given in Eqs.~\ref{eqn:updaterulest}--\ref{eqn:updateruleL}. When we incorporate the task descriptors into this framework, we alter   $\At{t}\rightarrow \ComboVec{t}$, $\bm{L}\rightarrow \ComboDict$, and $\bm{\Gamma}^{(t)}\rightarrow \bm{A}^{(t)}$. Note both $\ComboVec{t}$ and $\bm{A}^{(t)}$ are formed by adding deterministic entries and thus can be considered to be drawn i.i.d (because  $\bm{\Gamma}^{(t)}$ and $\At{t}$ are assumed to be drawn i.i.d). Therefore, incorporating task descriptors does not violate Condition 1. 

To show that Condition 2 holds, if we analogously form $\ComboDict_{\kappa}$, then the eigenvalues of  $\ComboDict_{\kappa}$ are strictly positive because they are either eigenvalues of $\bm{L}$ (which are strictly positive according to \cite{Ammar2014a}) or the regularizing parameter $\rho$ by definition. Thus, both conditions are met and convergence follows directly from Theorem~\ref{thrm:convergence}.

\subsection{Computational Complexity}
\label{sect:computationComplexity}

In this section, we analyze the computational complexity of TaDeLL.  Each update begins with one PG step to update $\bm{\alpha}^{(t)}$ and $\bm{\Gamma}^{(t)}$ at a cost of $O(\xi(d,n_t))$, where $\xi()$ depends on the base PG learner and $n_t$ is the number of trajectories obtained for task $\Task{t}$.  The cost of updating $\bm{L}$ and $\st{t}$ alone is $O(k^2d^3)$ \cite{Ruvolo2013}, and so the cost of updating $\ComboDict$ through coupled dictionary learning is $O(k^2(d+d_m)^3)$.  This yields an overall per-update cost of  $O(k^2(d+d_m)^3 + \xi(d,n_t))$, which is independent of $T$.

Next, we empirically demonstrate the benefits of TaDeLL on a variety of different learning problems.

\section{Evaluation on Reinforcement Learning Domains} \label{Experiments}
We apply TaDeLL to a series of RL problems. We consider the problem of learning a collection of different, related systems. For these systems, we use three benchmark control problems and an application to quadrotor stabilization.

\subsection{Benchmark Dynamical Systems}

\noindent \textbf{Spring Mass Damper (SM)}~~
The SM system is described by three parameters: the spring constant, mass, and damping constant. The system's state is given by the position and velocity of the mass. The controller applies a force to the mass, attempting to stabilize it to a given position. 

\noindent \textbf{Cart Pole (CP)}~~
The CP system involves balancing an inverted pendulum by applying a force to the cart. The system is characterized by the cart and pole masses, pole length, and a damping parameter. The states are the position and velocity of the cart and the angle and rotational velocity of the pole.

\noindent \textbf{Bicycle (BK)}~~
This system focuses on keeping a bicycle balanced upright as it rolls along a horizontal plane at constant velocity. The system is characterized by the bicycle mass, $x$- and $z$-coordinates of the center of mass, and parameters relating to the shape of the bike (the wheelbase, trail, and head angle). The state is the bike's tilt and its derivative; the actions are the torque applied to the handlebar and its derivative.

\begin{figure*}[tb!]
    \vspace{-10pt}
    \hspace{-10pt}
    \begin{subfigure}[b]{2.0in} %
       \includegraphics[clip,trim={100pt 240pt 120pt 250pt},height=1.5in]{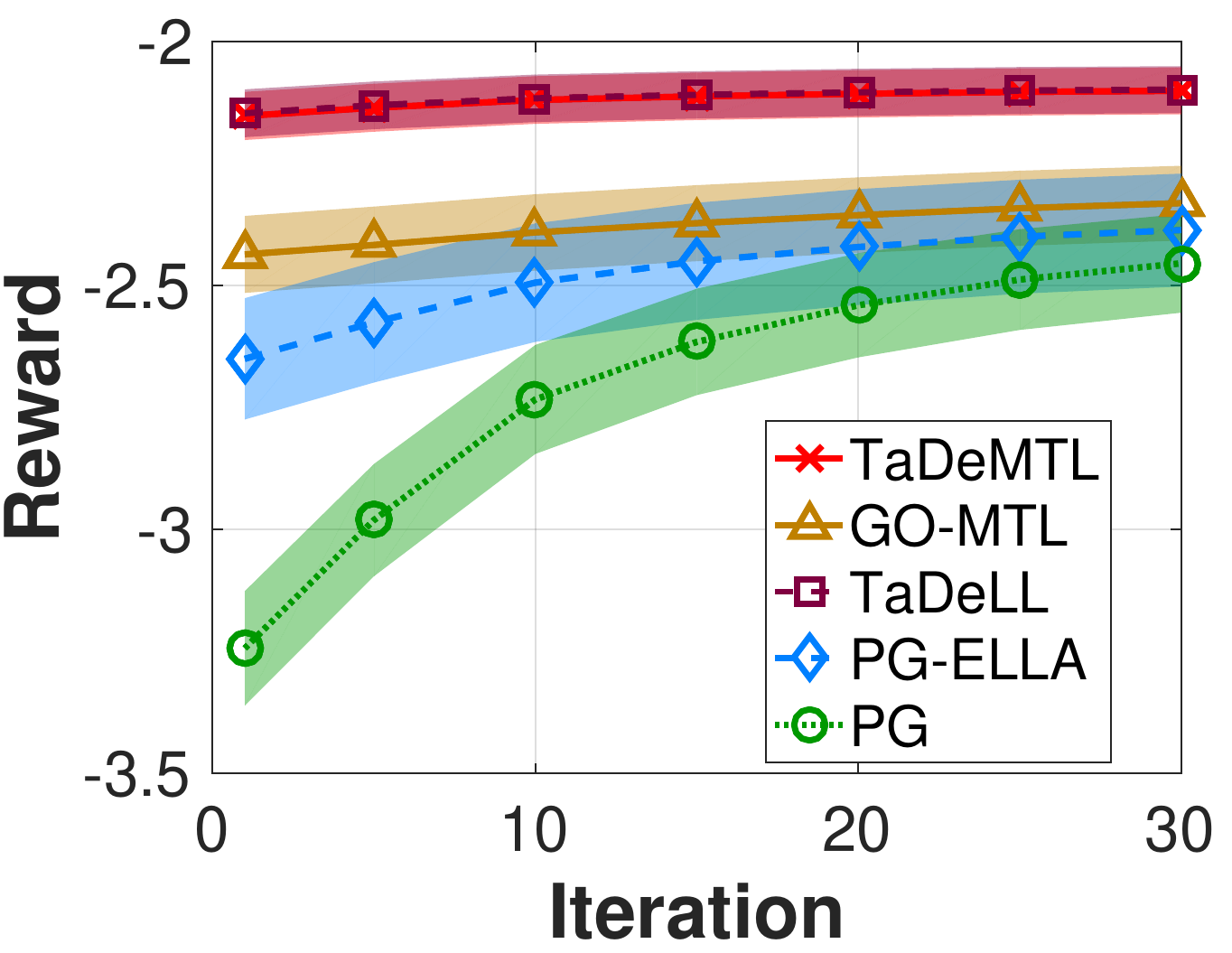}
        \caption{Simple Mass}
        \label{fig:SM_online}
    \end{subfigure}
    \begin{subfigure}[b]{1.9in}
    	\includegraphics[clip,trim={120pt 240pt 120pt 250pt},height=1.5in]{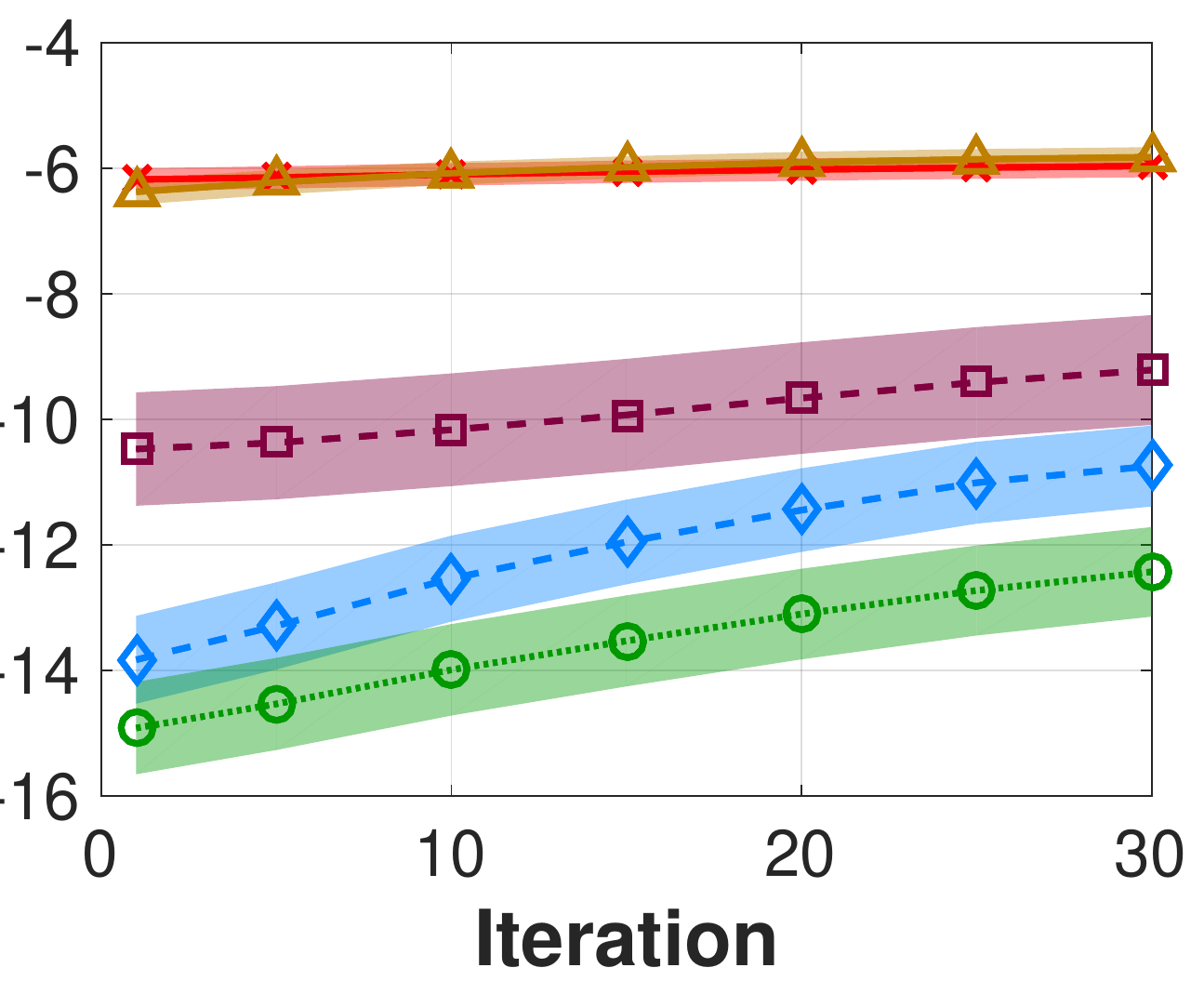}
        \caption{Cart Pole}
        \label{fig:CP_online}
    \end{subfigure}
    \begin{subfigure}[b]{1.9in}
    	\includegraphics[clip,trim={120pt 240pt 120pt 250pt},height=1.5in]{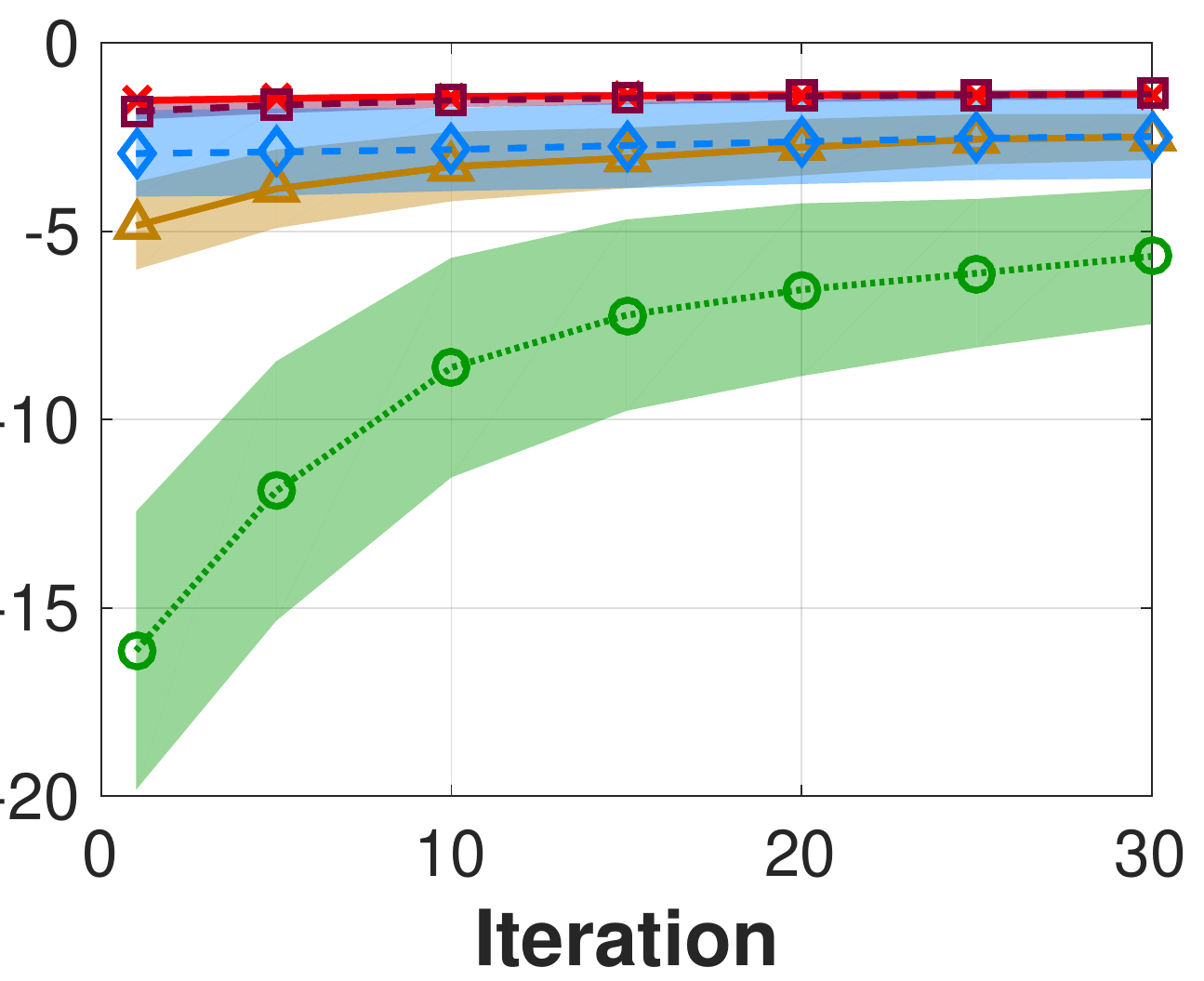}
        \caption{Bicycle}
        \label{fig:BK_online}
    \end{subfigure}
    
    \vspace{-0.5em}
    \caption{Performance of multi-task (solid lines), lifelong (dashed), and single-task learning (dotted) on benchmark dynamical systems. (Best viewed in color.)}\label{fig:results}
\end{figure*}

\subsection{Methodology} 
In each domain we generated 40 tasks, each with different dynamics, by varying the system parameters. The reward for each task was taken to be the distance between the current state and the goal.  For lifelong learning, tasks were encountered consecutively with repetition, and learning proceeded until each task had been seen at least once.  We used the same random task order between methods to ensure fair comparison.  The learners sampled trajectories of 100 steps, and the learning session during each task presentation was limited to 30 iterations.     For MTL, all tasks weres presented simultaneously.  We used Natural Actor Critic ~\cite{peters2008natural} as the base learner for the benchmark systems and episodic REINFORCE \cite{williams1992simple} for quadrotor control.  
We chose $k$ and the regularization parameters independently for each domain to optimize the combined performance of all methods on 20 held-out tasks, and set $\rho = \mathit{mean}(diag(\rho^{(t)}))$ to balance the fit to the descriptors and the policies. We measured learning curves based on the final policies for each of the 40 tasks. %
The system parameters for each task were used as the task descriptor features $\phi(\bm{m})$; we also tried several non-linear transformations as $\phi(\cdot)$, but found the linear features worked well.  %

\subsection{Results on Benchmark Systems}

\begin{figure*}[tb!]
    \centering
    \vspace{-10pt}
    \hspace{-10pt}
           \begin{subfigure}[b]{2in}
       \includegraphics[clip,trim={100pt 250pt 150pt 250pt},height=1.5in]{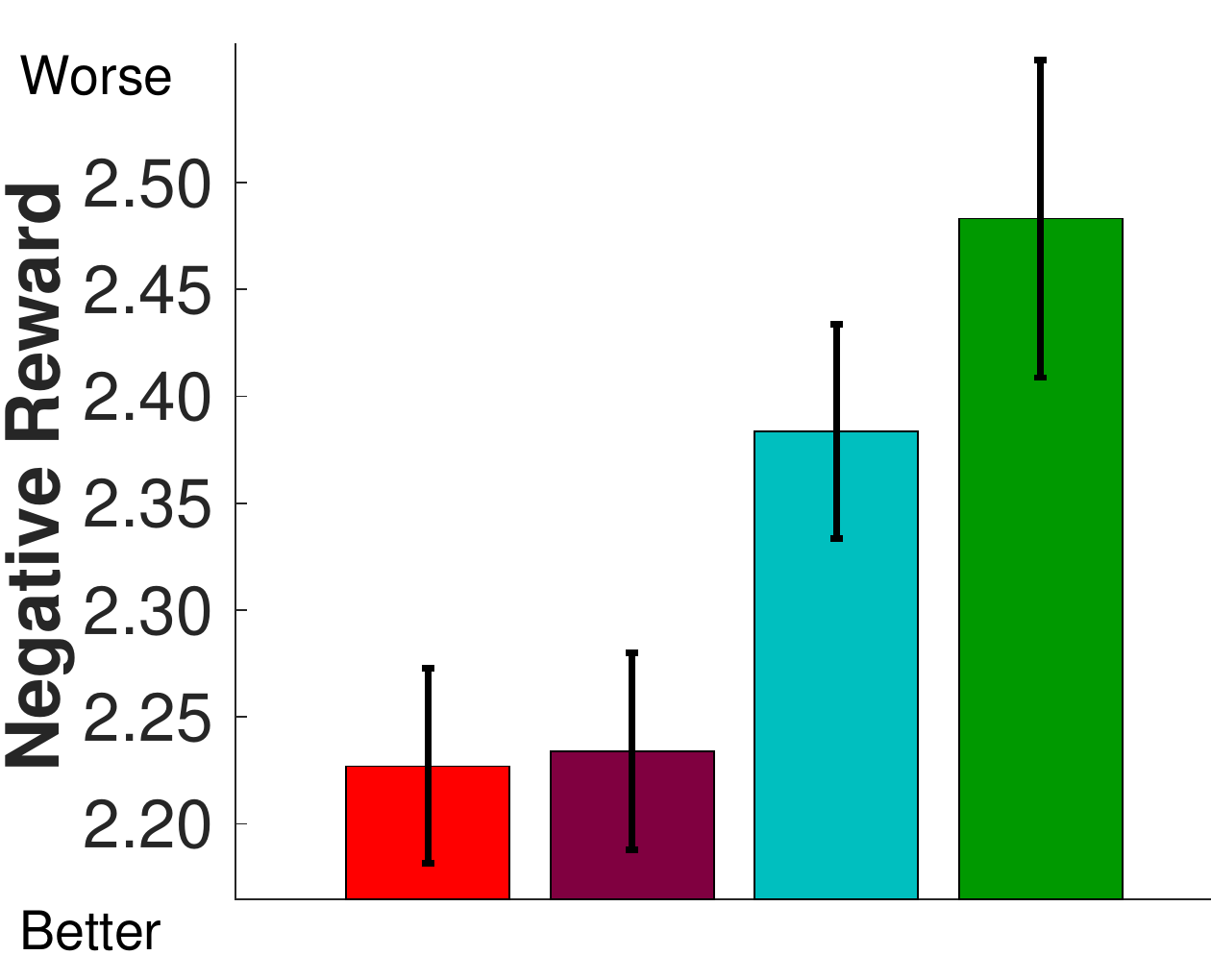}
        \caption{Simple Mass}
        \label{fig:SM_zero}
    \end{subfigure}
    \begin{subfigure}[b]{1.9in}
    	\includegraphics[clip,trim={150pt 250pt 150pt 250pt},height=1.5in]{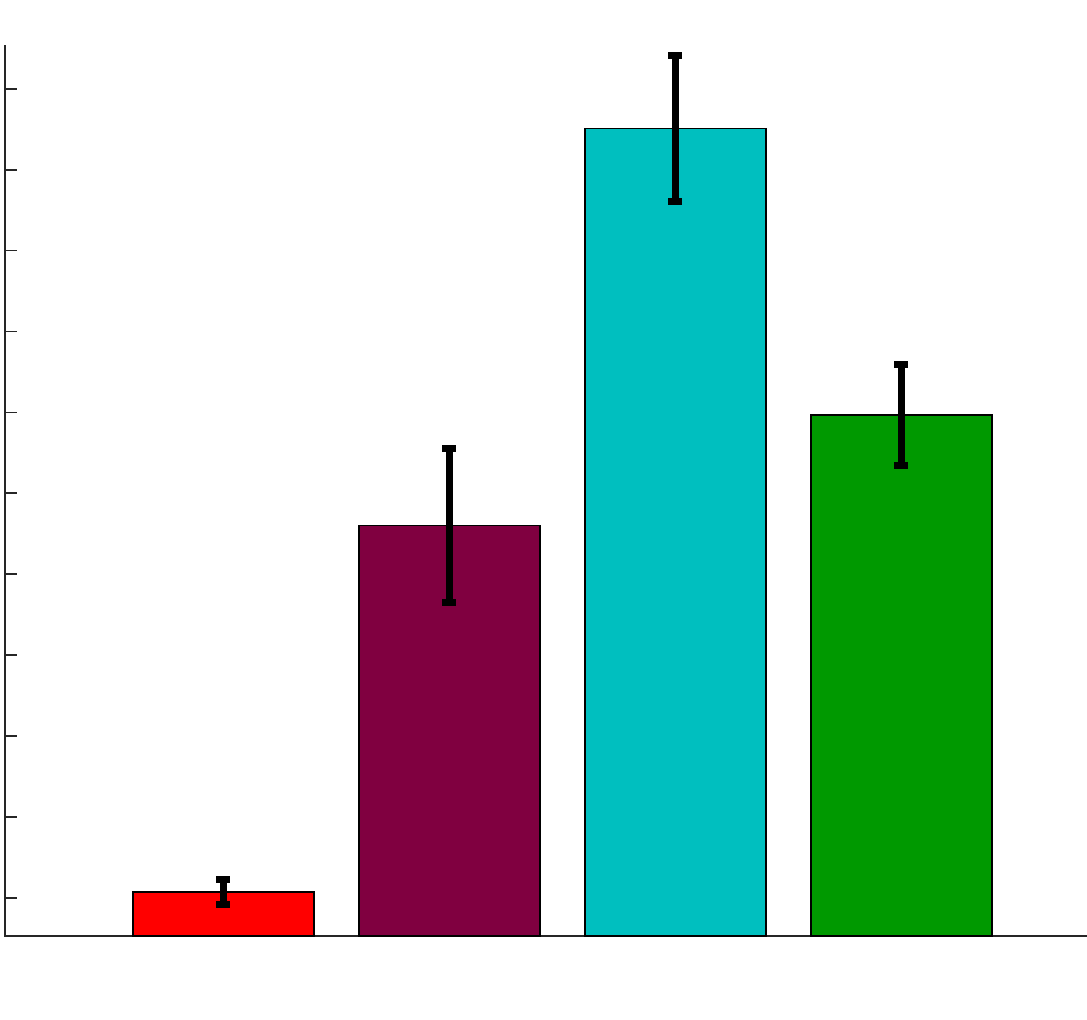}
        \caption{Cart Pole}
        \label{fig:CP_zero}
    \end{subfigure}
    \begin{subfigure}[b]{1.9in}
    	\includegraphics[clip,trim={150pt 250pt 150pt 250pt},height=1.5in]{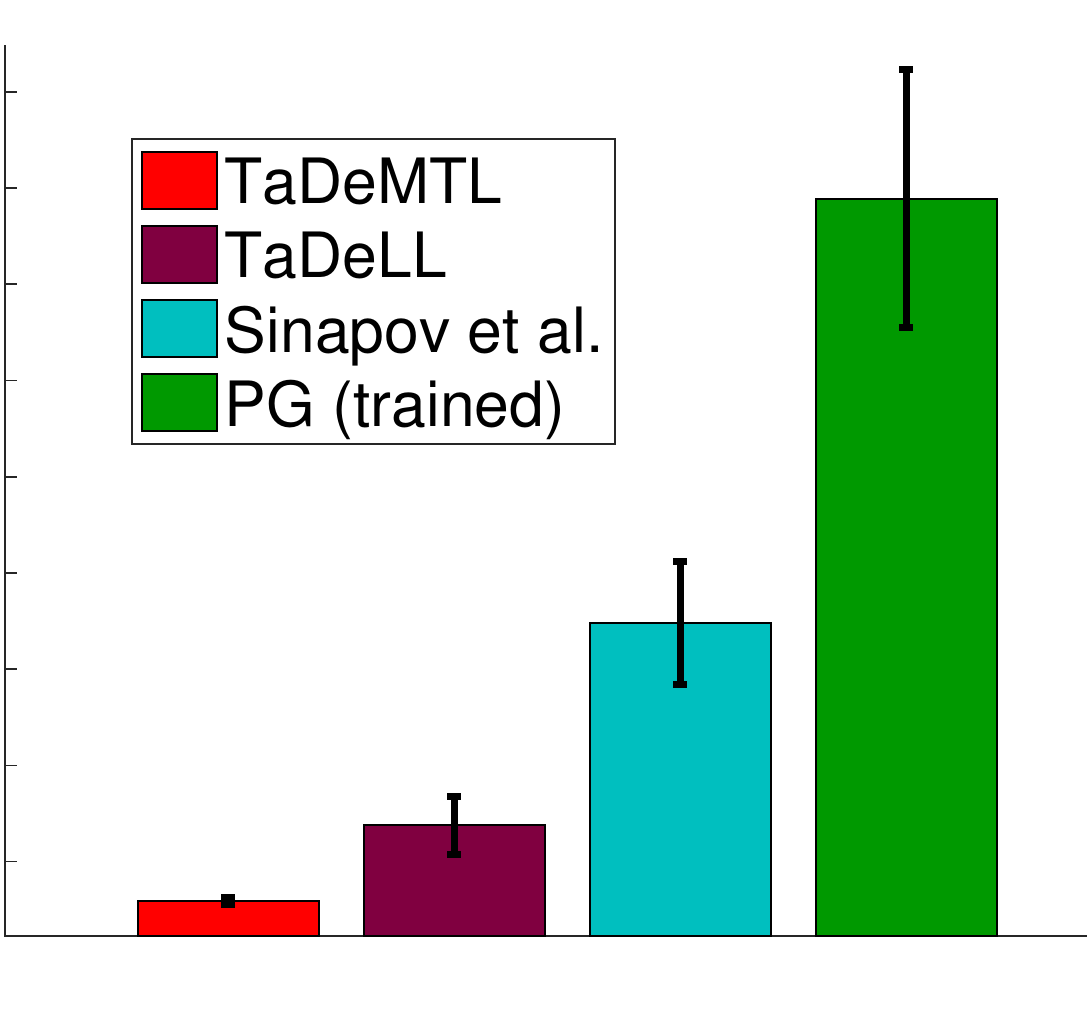}
        \caption{Bicycle}
        \label{fig:BK_zero}
    \end{subfigure}
    \vspace{-0.25em}
     \caption{Zero-shot transfer to new tasks. The figure shows the initial ``jumpstart'' improvement  on each task domain. (Best viewed in color.)}\label{fig:zeroshotJumpstart}
\end{figure*}

\begin{figure*}[tb!]
    \centering
    \vspace{-10pt}
    \hspace{-10pt}
           \begin{subfigure}[b]{2in} 
       \includegraphics[clip,trim={100pt 240pt 120pt 250pt},height=1.5in]{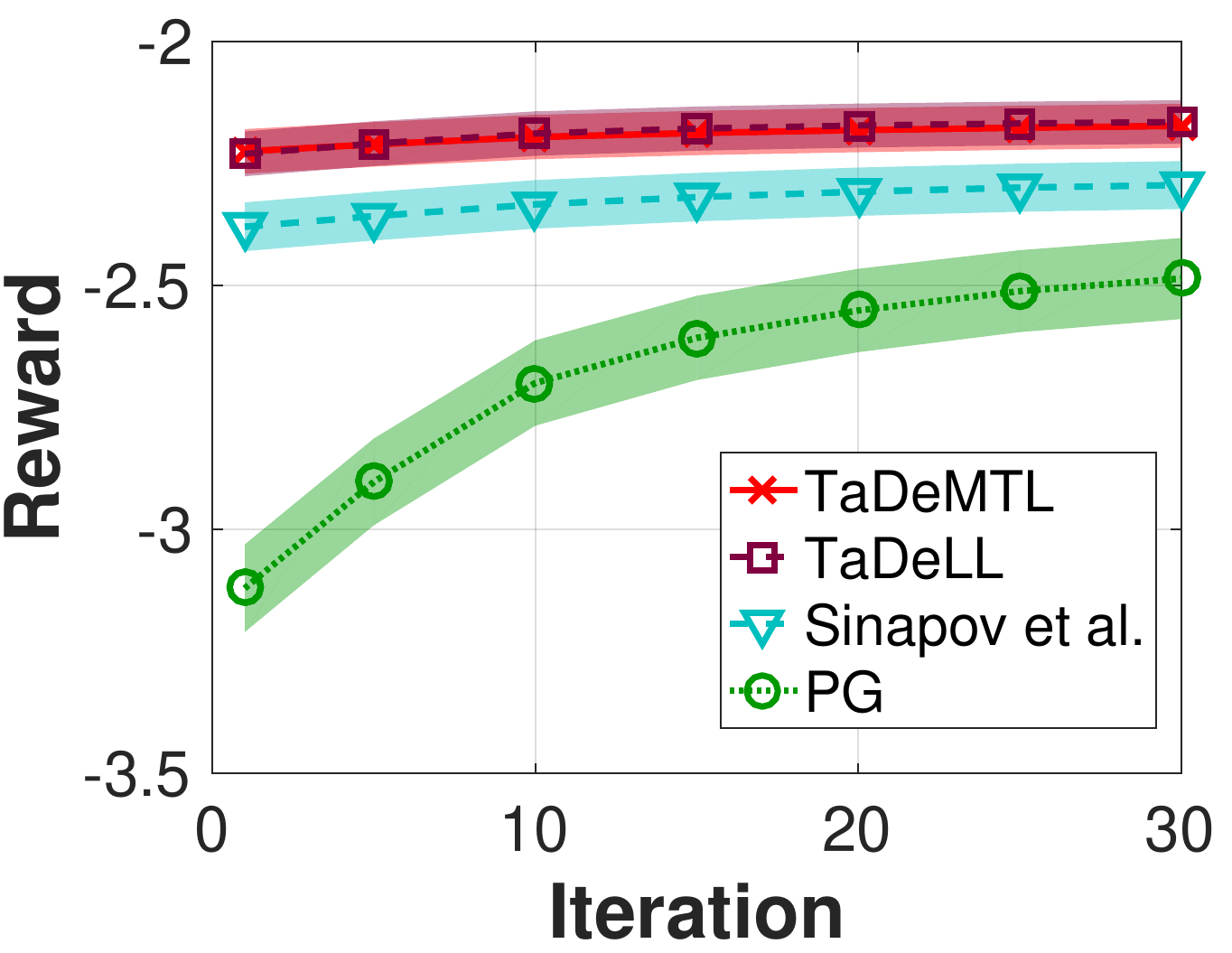}
        \caption{Simple Mass}
        \label{fig:SM_jump}
    \end{subfigure}
    \begin{subfigure}[b]{1.9in}
    	\includegraphics[clip,trim={100pt 240pt 120pt 250pt},height=1.5in]{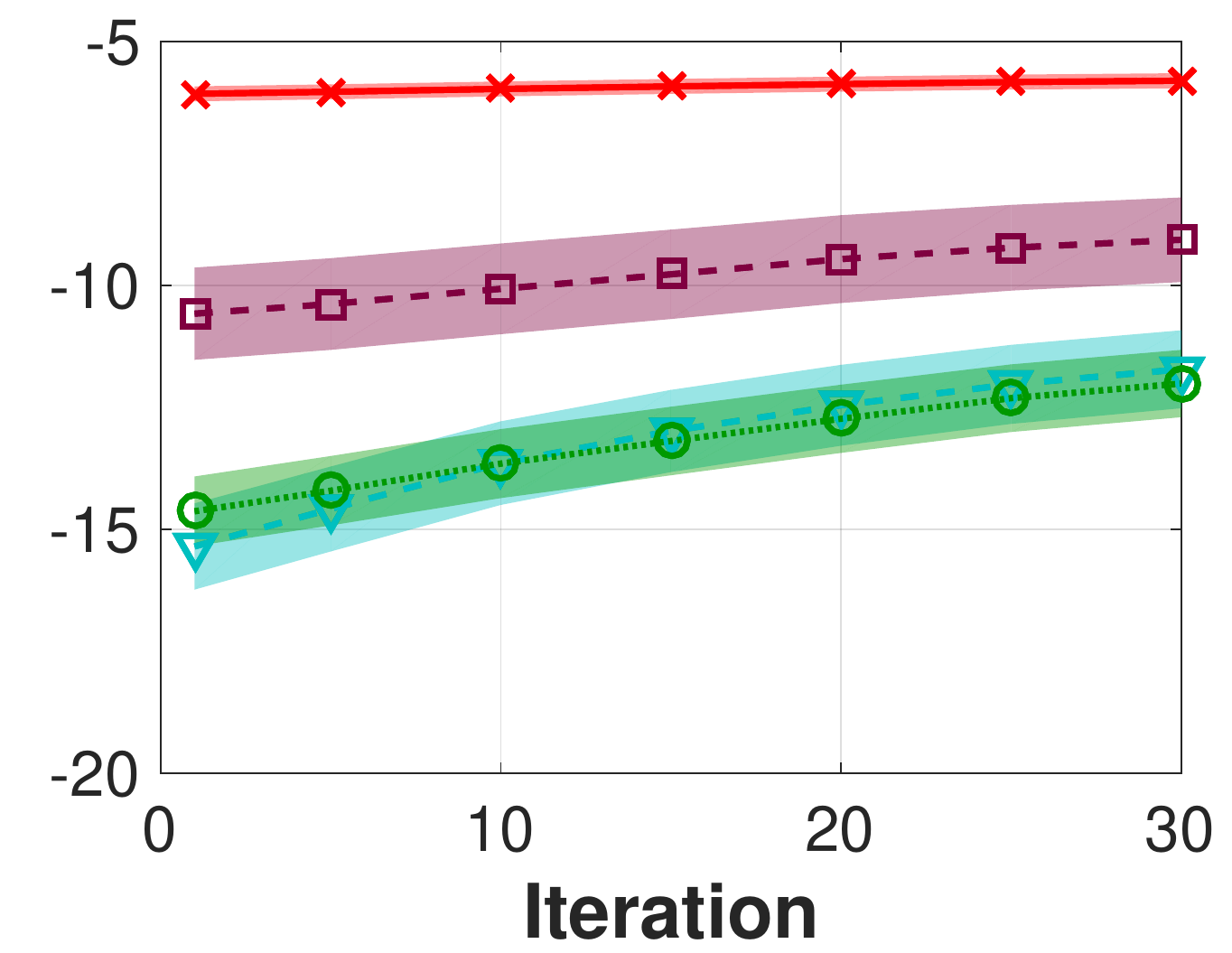}
        \caption{Cart Pole}
        \label{fig:CP_jump}
    \end{subfigure}
    \begin{subfigure}[b]{1.9in}
    	\includegraphics[clip,trim={100pt 240pt 120pt 250pt},height=1.5in]{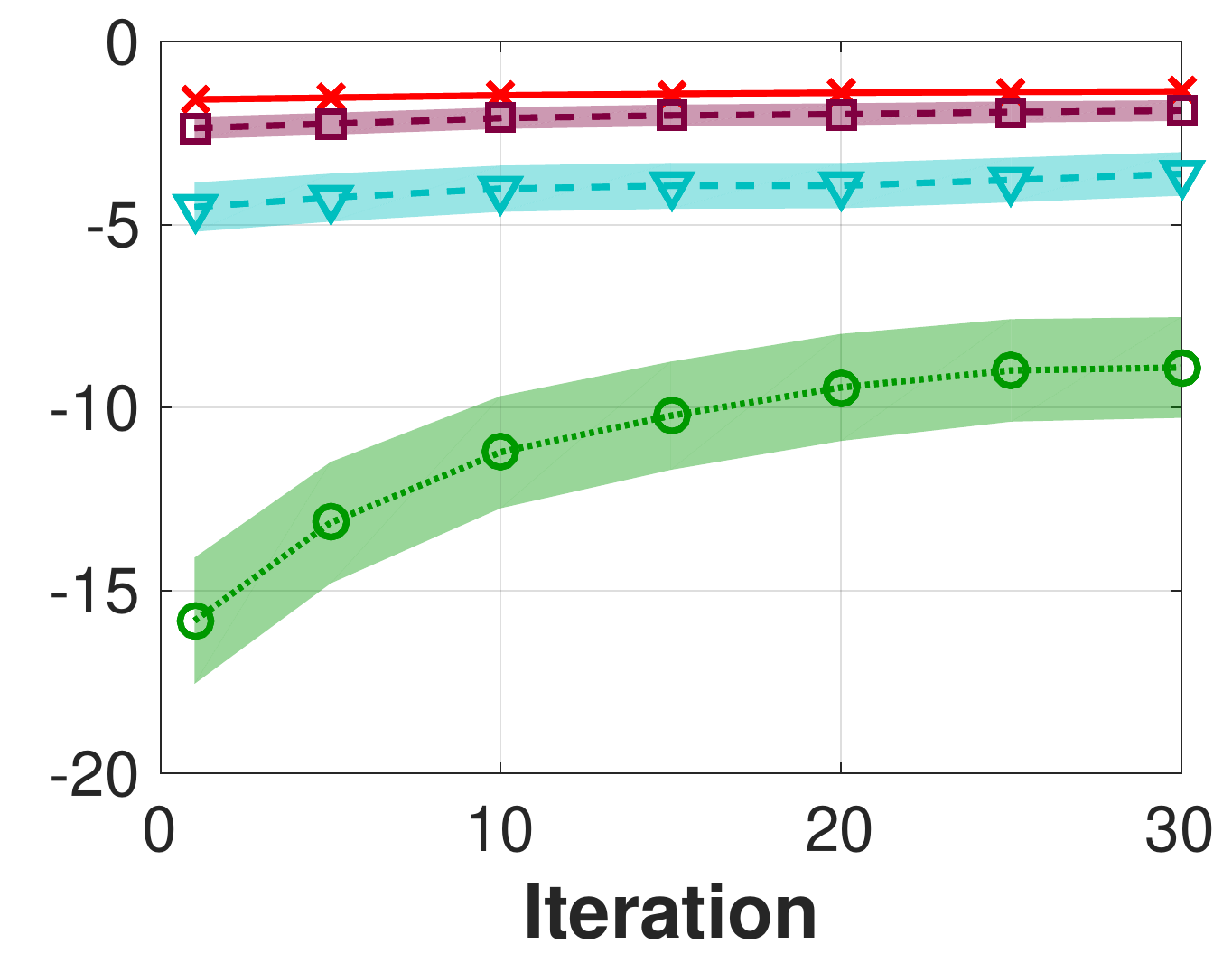}
        \caption{Bicycle}
        \label{fig:BK_jump}
    \end{subfigure}
        \caption{Learning performance of using the zero-shot policies as warm start initializations for PG. The performance of the single-task PG learner is included for comparison. (Best viewed in color.)}\label{fig:zeroshotWarmstart} 
\end{figure*}

Figure~\ref{fig:results} compares our TaDeLL approach for lifelong learning with task descriptors to 1.)~PG-ELLA~\cite{Ammar2014a}, which does not use task features, 2.)~GO-MTL~\cite{Kumar2012}, the MTL optimization of Eq.~\ref{eqn:MTLObjective}, and 3.)~single-task learning using PG.  For comparison, we also performed an offline MTL optimization of Eq.~\ref{eqn:MTLObjectiveCoupledDictionaries} via alternating optimization, and plot the results as TaDeMTL. The shaded regions on the plots denote standard error bars.

We see that task descriptors improve lifelong learning on every system, even driving performance to a level that is unachievable from training the policies from experience alone via GO-MTL in the SM and BK domains.  The difference between TaDeLL and TaDeMTL is also negligible for all domains except CP, demonstrating the effectiveness of our online optimization.

To measure zero-shot performance, we generated an additional 40 tasks for each domain, averaging results over these new tasks.  Figure~\ref{fig:zeroshotJumpstart} shows that task descriptors are effective for zero-shot transfer to new tasks.  We see that our approach improves the initial performance  (i.e., the ``jumpstart'' \cite{taylor2009transfer}) on new tasks, outperforming Sinapov~\etal~\citeyear{Sinapov2015}'s method  and single-task PG, which was allowed to train on the task.   We attribute the especially poor performance of Sinapov~\etal~on CP to the fact that the CP policies differ substantially; in domains where the source policies are vastly different from the target policies, Sinapov~\etal's algorithm does not have an appropriate source to transfer.  Their approach is also much more computationally expensive (quadratic in the number of tasks) than our approach (linear in the number of tasks), as shown in Figure~\ref{fig:time}; details of the runtime experiments are included in Section~\ref{computation}.  Figure~\ref{fig:zeroshotWarmstart} shows that the zero-shot policies can be used effectively as a warm start initialization for a PG learner, which is then allowed to improve the policy.

\subsection{Application to Quadrotor Control}

We also applied our approach to the more challenging domain of quadrotor control, focusing on zero-shot transfer to new stability tasks. To ensure realistic dynamics, we use the model of Bouabdallah and Siegwart~\citeyear{bouabdallah2005backstepping}, which has been verified on physical systems. The quadrotors are characterized by three inertial constants and the arm length, with their state consisting of roll/pitch/yaw and their derivatives.

\begin{figure}[tb!]%
    \centering
        \includegraphics[trim={150pt 250pt 150pt 250pt}, height=.4\textwidth]{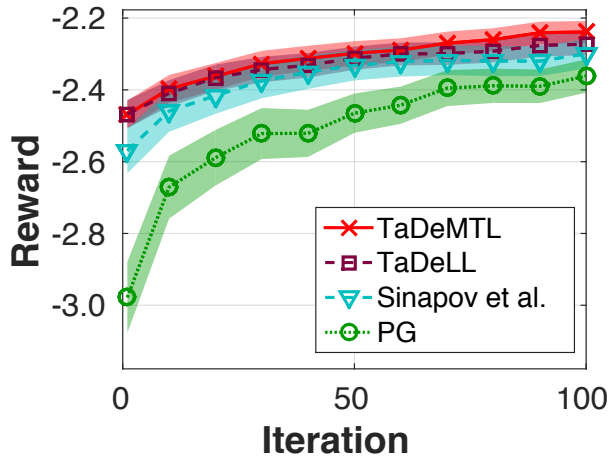}      
        \vspace{-0.25em}
        \caption{Warm start learning on quadrotor control. (Best viewed in color.)}
        \label{fig:Quad_zero}
\end{figure}

Figure \ref{fig:Quad_zero} shows the results of our application, demonstrating that TaDeLL can predict a controller for new quadrotors through zero-shot learning that has equivalent accuracy to PG, which had to train on the system.  As with the benchmarks, TaDeLL is effective for warm start learning with PG.

\section{Evaluation on Supervised Learning Domains} 
\label{ClassificationRegression}

In this section, we evaluate TaDeLL on regression and classification domains, considering the problem of predicting the real-valued location of a robot's end effector and two synthetic classification tasks.

\subsection{Predicting the Location of a Robot End Effector}

In this section, we evaluate TaDeLL on a regression domain.  We look at the problem of predicting the real-valued position of the end effector of an 8-DOF robotic arm in 3D space, given the angles of the robot joints. %
Different robots have different link lengths, offsets, and twists, and we use these parameters as the description of the task. %

We consider 200 different robot arms and use 10 points as training data per robot. The robot arms are simulated using the Robot Toolbox \cite{Corkella}. The learned dictionaries are then used to predict models for 200 different unseen robots. We measure performance as the mean square error of the prediction against the true location of the end effector.

Table~\ref{tab:robot} shows that both TaDeLL and ELLA outperform the single-task learner, with TaDeLL slightly outperforming ELLA. This same improvement holds for zero-shot prediction on new robot arms, with TaDeLL outperforming the single-task learner, which was trained on the new robot. %

\begin{figure}[t!]
    \centering
        \begin{subfigure}[b]{0.45\textwidth}
    	\includegraphics[clip,trim={100pt 230pt 100pt 230pt}, width=\textwidth]{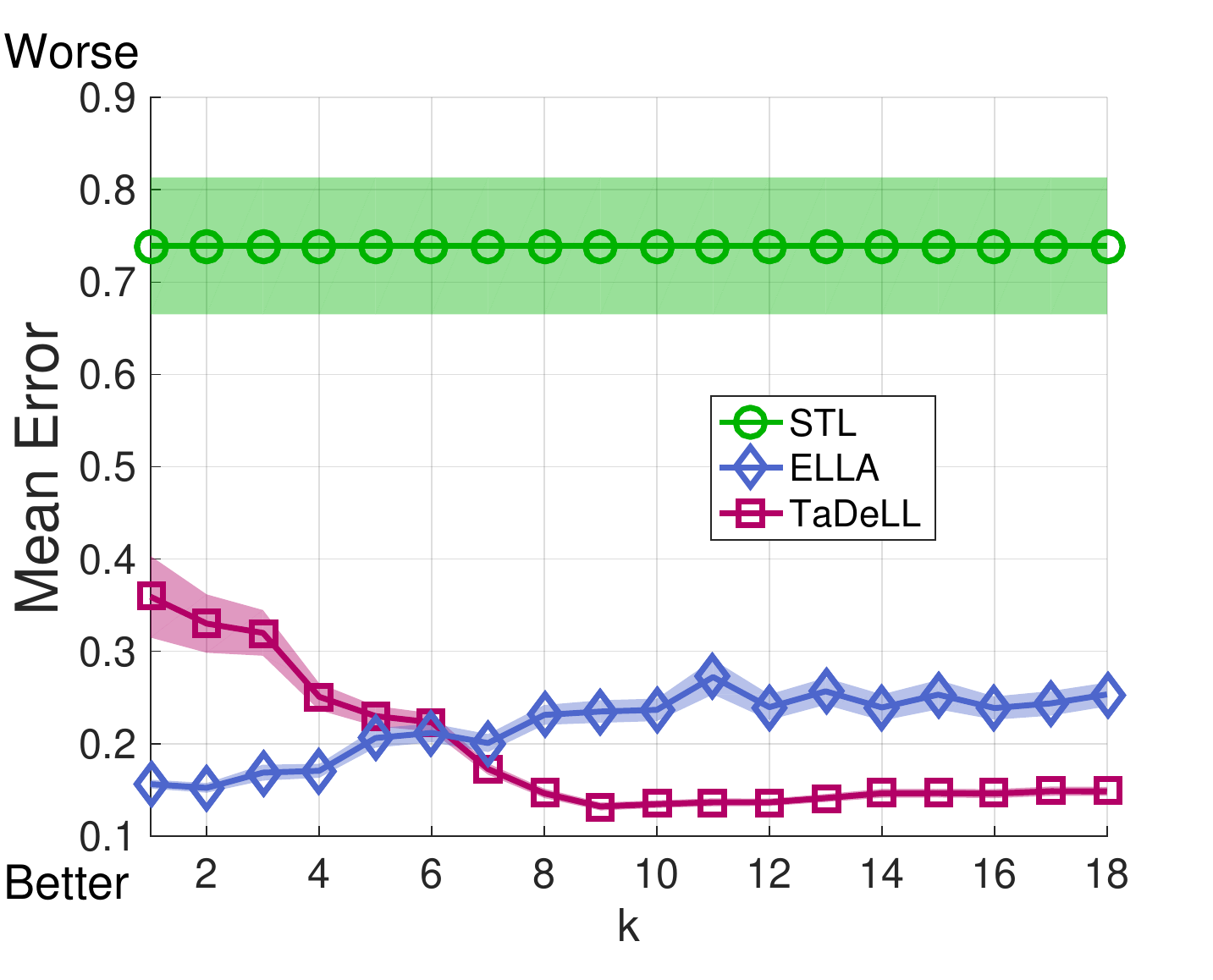}
	\caption{Lifelong Learning.}
	\end{subfigure}
	\begin{subfigure}[b]{0.45\textwidth}
	\includegraphics[clip,trim={100pt 230pt 100pt 230pt}, width=\textwidth]{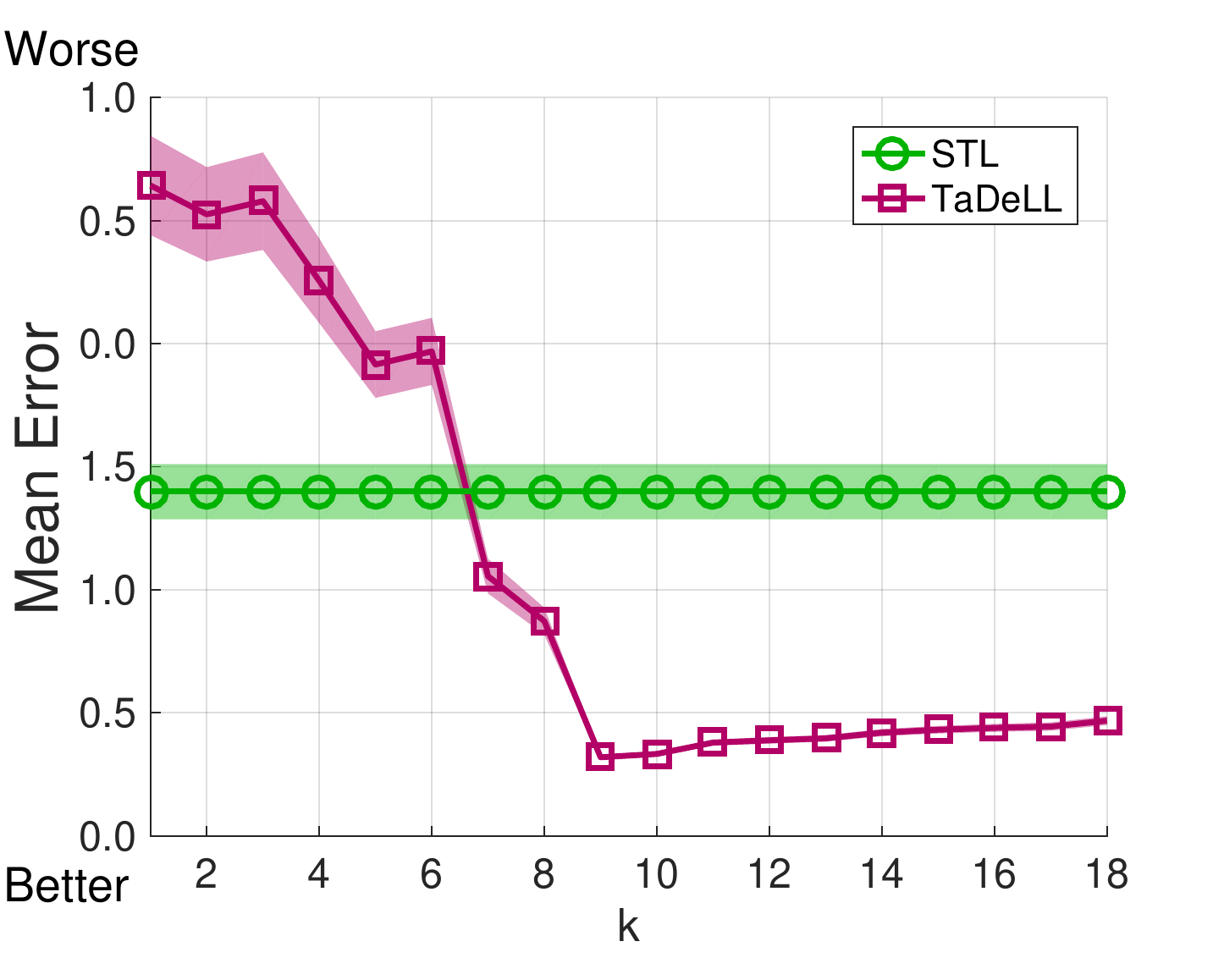}
	\caption{Zero-shot Prediction.}
	\end{subfigure}
        \caption{Performance of TaDeLL and ELLA as the dictionary size $k$ is varied for lifelong learning and zero-shot learning. Performance of the single task learner is provided for comparison. In the lifelong learning setting, both TaDeLL and ELLA demonstrate positive transfer that converges to the performance of the single task learner as $k$ is increased. We see that, for this problem, TaDeLL prefers a slightly larger value of $k$.}\label{fig:varyk}
\end{figure}

\begin{figure}[t!]
    \centering
    	\includegraphics[clip,trim={100pt 230pt 100pt 230pt}, width=.6\textwidth]{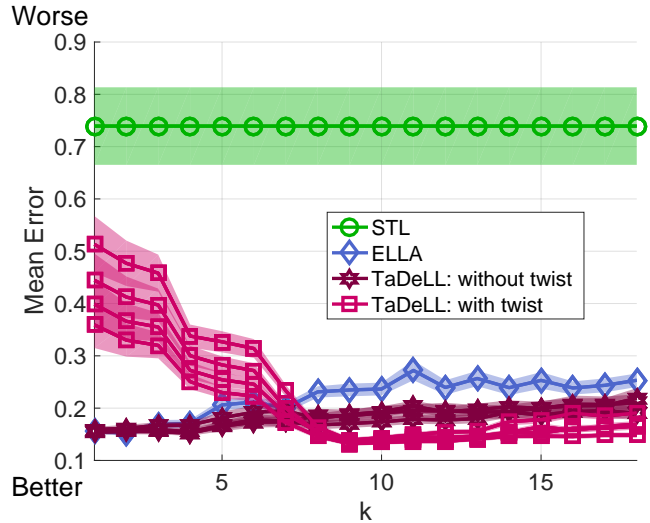}
        \caption{An ablative experiment studying the performance of TaDeLL as a function of the dictionary size $k$, as we vary the subset of descriptors used. The feature consist of twist(t), length(l), and offset(o) variables for each joint. We train TaDeLL using only subsets of the features $\{t,l,o,tl,to,lo,tlo\}$ and we see that the need for a larger $k$ is directly related to learning the \emph{twist}. Subsets that contain twist descriptors are shown in magenta. Trials that do not include twist descriptors are shown in gray. Performance of ELLA and the single-task learner (STL) are provided for comparison. (Best viewed in color.)}\label{fig:varyk_tlo}
\end{figure}

To better understand the relationship of dictionary size to performance, we investigated how learning performance varies with the number of bases $k$ in the dictionary.  Figure \ref{fig:varyk} shows this relationship for the lifelong learning and zero-shot prediction settings. We observe that TaDeLL performs better with a larger dictionary than ELLA, we hypothesize that difference results from the added difficulty of encoding the representations with the task descriptions. To test this hypothesis, we reduced the number of descriptors in an ablative experiment. Recall that the task has 24 descriptors consisting of a twist, link offset, and link length for each joint. We reduced the number of descriptors by alternatingly removing the subsets of features corresponding to the twist, offset, and length. Figure \ref{fig:varyk_tlo} shows the performance of this ablative experiment, revealing that the need for the increased number of bases is particularly related to learning \emph{twist}.  %

\begin{table} 
\centering
\begin{tabular}{l|c|c}
Algorithm & Lifelong Learning & Zero-Shot Prediction \\\hline
TaDeLL & 0.131 $\pm$ 0.004 & 0.159 $\pm$ 0.005\\
ELLA & 0.152 $\pm$ 0.005 & N/A\\
STL & 0.73 $\pm$ 0.07 & 0.70 $\pm$ 0.05
\end{tabular}
\caption{Regression performance on robot end effector prediction in both lifelong learning and zero-shot settings.  Performance is measured in mean squared error.}
\label{tab:robot}
\end{table}

\begin{figure}[t!]
    \centering
    \hspace{1in}
    \begin{subfigure}[b]{3in}\hspace{1in}
    \includegraphics[height=2in]{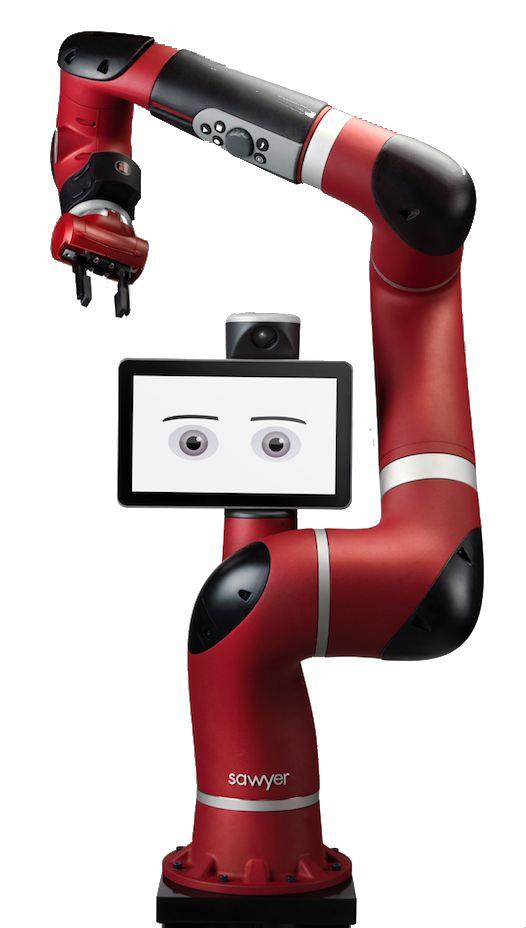}
    \end{subfigure}~ \hspace{-0.75in}
    \begin{subfigure}[b]{3in}%
    	\includegraphics[clip,trim={160pt 265pt 200pt 265pt}, height=.75\textwidth]{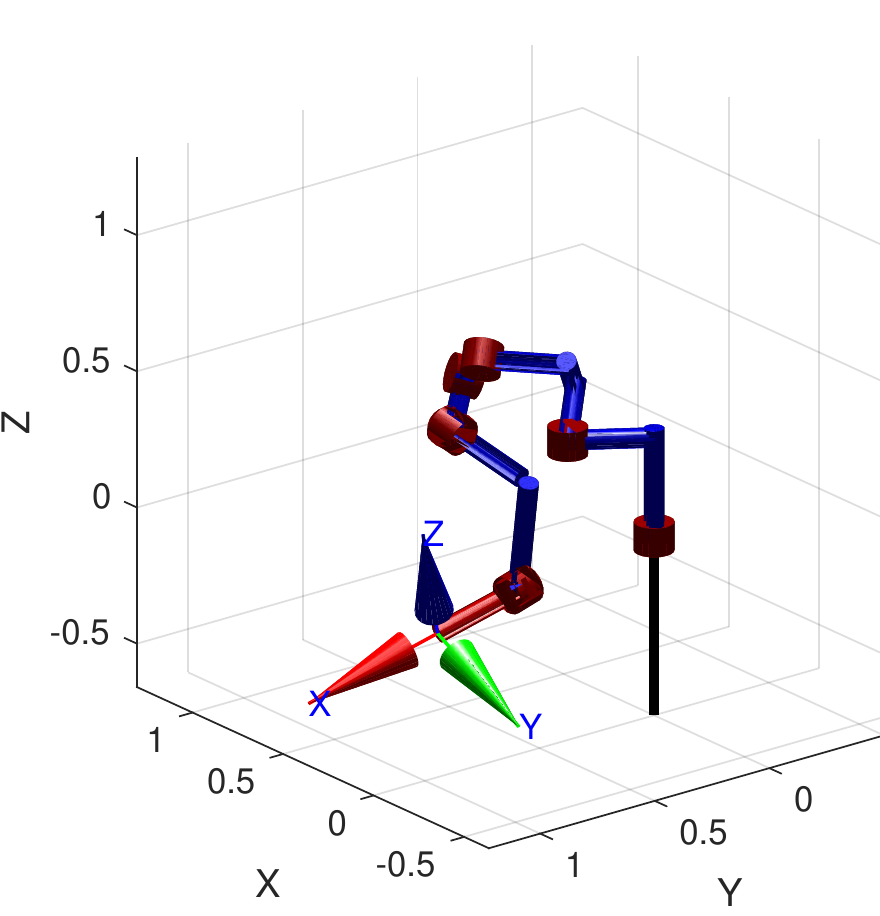}
        \end{subfigure}
        \label{fig:numfeat}
         \caption{Example model of an 8-DOF robot. (Photo of the Sawyer arm by Rethink Robotics.)}
\end{figure}

\subsection{Experiments on Synthetic Classification Domains}

To better understand the connections between TaDeLL's performance and the structure of the tasks, we evaluated TaDeLL on two synthetic classification domains.  The use of synthetic domains allows us to tightly control the task generation process and the relationship between the target model and the descriptor.

The first synthetic domain consists of binary-labeled instances drawn from $\Reals^8$, and each sample $\bm{x}$ belongs to the positive class iff $\bm{x}^\transpose \bm{m} > 0$. Each task has a different parameter vector $\bm{m}$ drawn from the uniform distribution $\bm{m} \in [-0.5,0.5]$; these vectors $\bm{m}$ are also used as the task descriptors. Note that by sampling $\bm{m}$ from the uniform distribution, this domain violates the assumptions of ELLA that the samples are drawn from a common  set of latent features.  Each task's data consists of 10 training samples, and we generated 100 tasks to evaluate lifelong learning.

Table \ref{tab:synth1} shows the performance on this Synthetic Domain 1.  We see that the inclusion of meaningful task descriptors enables TaDeLL to learn a better dictionary than ELLA in a lifelong learning setting. We also generated an additional 100 unseen tasks to evaluate zero-shot prediction, which is similarly successful. %

\begin{table} 
\centering
\begin{tabular}{l|c|c}
Algorithm & Lifelong Learning & Zero-Shot Prediction\\\hline
TaDeLL & 0.926 $\pm$ 0.004  & 0.930 $\pm$ 0.002 \\
ELLA & 0.814 $\pm$ 0.008 & N/A\\
STL & 0.755 $\pm$ 0.009 & 0.762 $\pm$ 0.008
\end{tabular}
\caption{Classification accuracy on Synthetic Domain 1.}
\label{tab:synth1}
\end{table}

For the second synthetic domain, we generated $\bm{L}$ and $\bm{D}$ matrices, and then generated a random sparse vector $\bm{s}^{(t)}$ for each task.  The true task model is then given by a logistic regression classifier with $\bm{\theta}^{(t)} = \bm{L}\bm{s}^{(t)}$. This generation process directly follows the assumptions of ELLA and TaDeLL, where $\bm{D}$ is generated independently.\comment{ and can be seen as a co-view of $\bm{L}$ according to the co-training literature~\cite{blum1998combining}.} We similarly generate 100 tasks for lifelong learning and another 100 unseen tasks for zero-shot prediction, and use the true task models to label 10 training points per task.  In this experiment, we empirically demonstrate that TaDeLL works in the case of this assumption (Table~\ref{tab:synth2}) in both lifelong learning and zero-shot prediction settings.

\begin{figure}[t!]
    \centering %
    \begin{subfigure}[b]{3in}
        \includegraphics[clip,trim={100pt 200pt 100pt 200pt},height=2.75in]{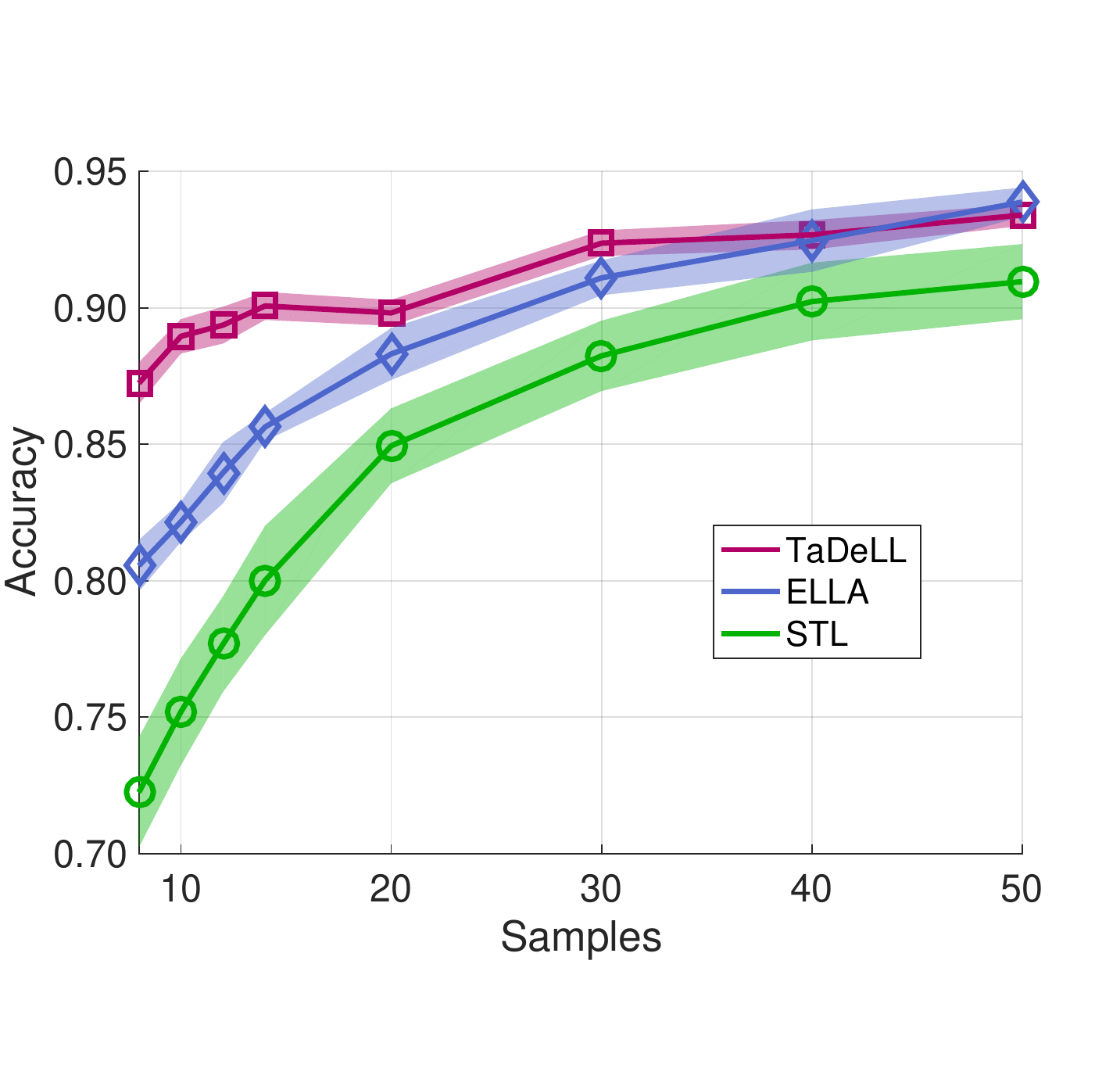}
        \caption{Lifelong Learning}
        \label{fig:samp}
    \end{subfigure}~ %
    \begin{subfigure}[b]{3in}
    	\includegraphics[clip,trim={100pt 200pt 100pt 200pt},height=2.75in]{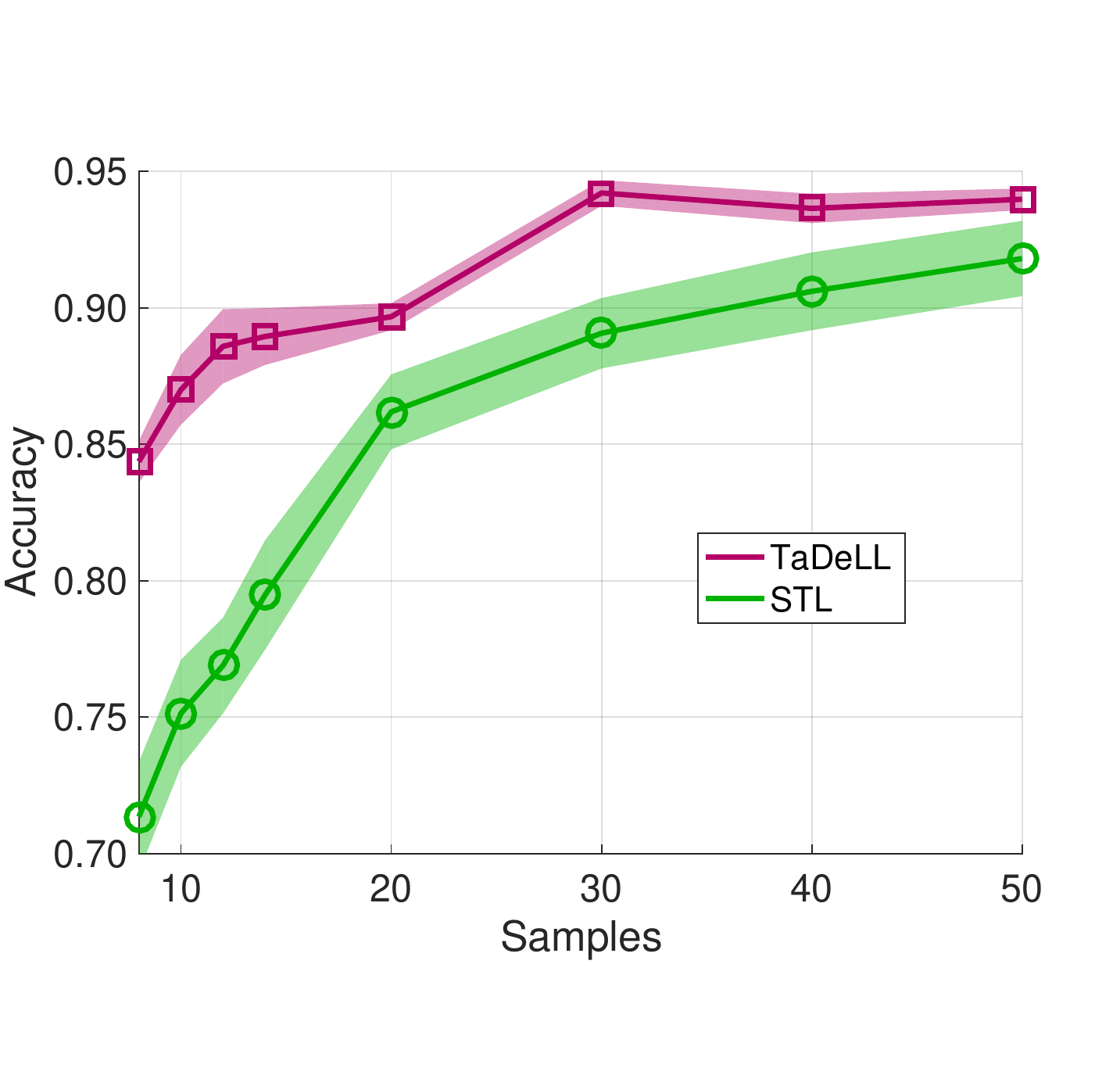}
        \caption{Zero-Shot Prediction}
        \label{fig:samp_zero}
    \end{subfigure}
    	\caption{Performance versus sample complexity on Synthetic Domain 2.}
\end{figure}

We also use this domain to investigate performance versus sample complexity, as we generated varying amounts of training data per task. In Figure \ref{fig:samp}, we see that TaDeLL is able to greatly improve performance given on a small number of samples, and as expected, its benefit becomes less dramatic as the single-task learner receives sufficient samples. Figure~\ref{fig:samp_zero} shows similar behavior in the zero-shot case.

\begin{table} 
\centering
\begin{tabular}{l|c|c}
Algorithm & Lifelong Learning & Zero-Shot Prediction \\\hline
TaDeLL & 0.889 $\pm$ 0.006 & 0.87 $\pm$ 0.01\\
ELLA & 0.821 $\pm$ 0.007 & N/A\\
STL & 0.752 $\pm$ 0.009 & 0.751 $\pm$ 0.009
\end{tabular}
\caption{Classification accuracy on Synthetic Domain 2.}
\label{tab:synth2}
\end{table}

\section{Additional Experiments}

Having shown how TaDeLL can improve learning in a variety of settings, we now turn our attention to understanding other aspects of the algorithm. Specifically, we look at the issue of task descriptor selection and partial information, runtime comparisons, and the effect of varying the number of tasks used to train the dictionaries.  

\subsection{Choice of Task Descriptor Features} \label{whichfeatures}

For RL, we used the system parameters as the task description, and for the robot end effector prediction, we used the dimensions of the robot. While in these cases the choice of task descriptor was straightforward, this might not always be the case. It is unclear exactly how the choice of task descriptor features might affect the resulting performance. In other scenarios, we may have only partial knowledge of the system parameters. 

\begin{figure}[t!]
    \centering
    \begin{subfigure}[b]{3in}
        \includegraphics[clip,trim={80pt 250pt 100pt 250pt},height=2.0in]{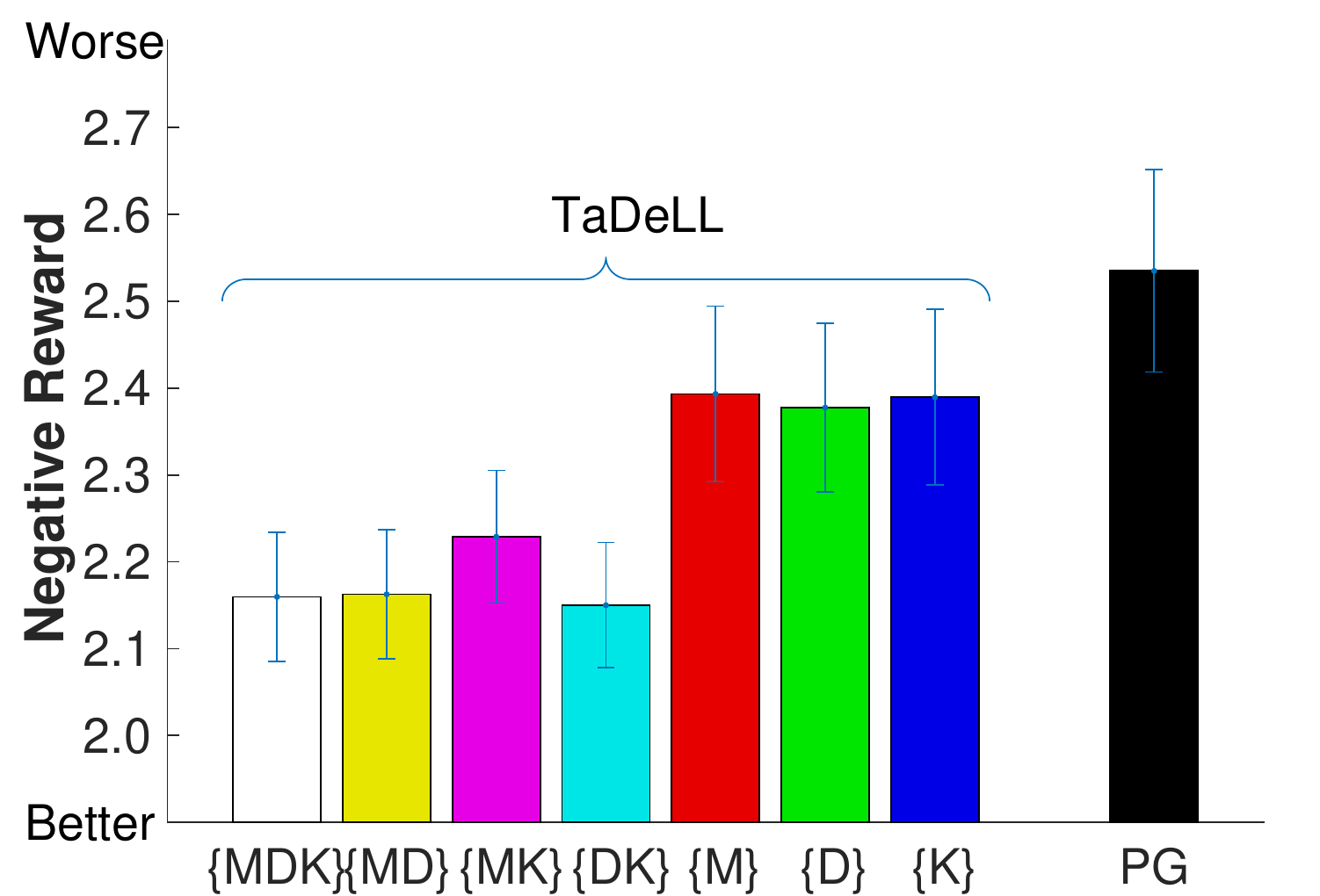}
        \caption{Spring-Mass RL}
        \label{fig:sm_numfeatures}
    \end{subfigure}~
    \begin{subfigure}[b]{3in}
    	\includegraphics[clip,trim={40pt 200pt 90pt 225pt},height=1.8in]{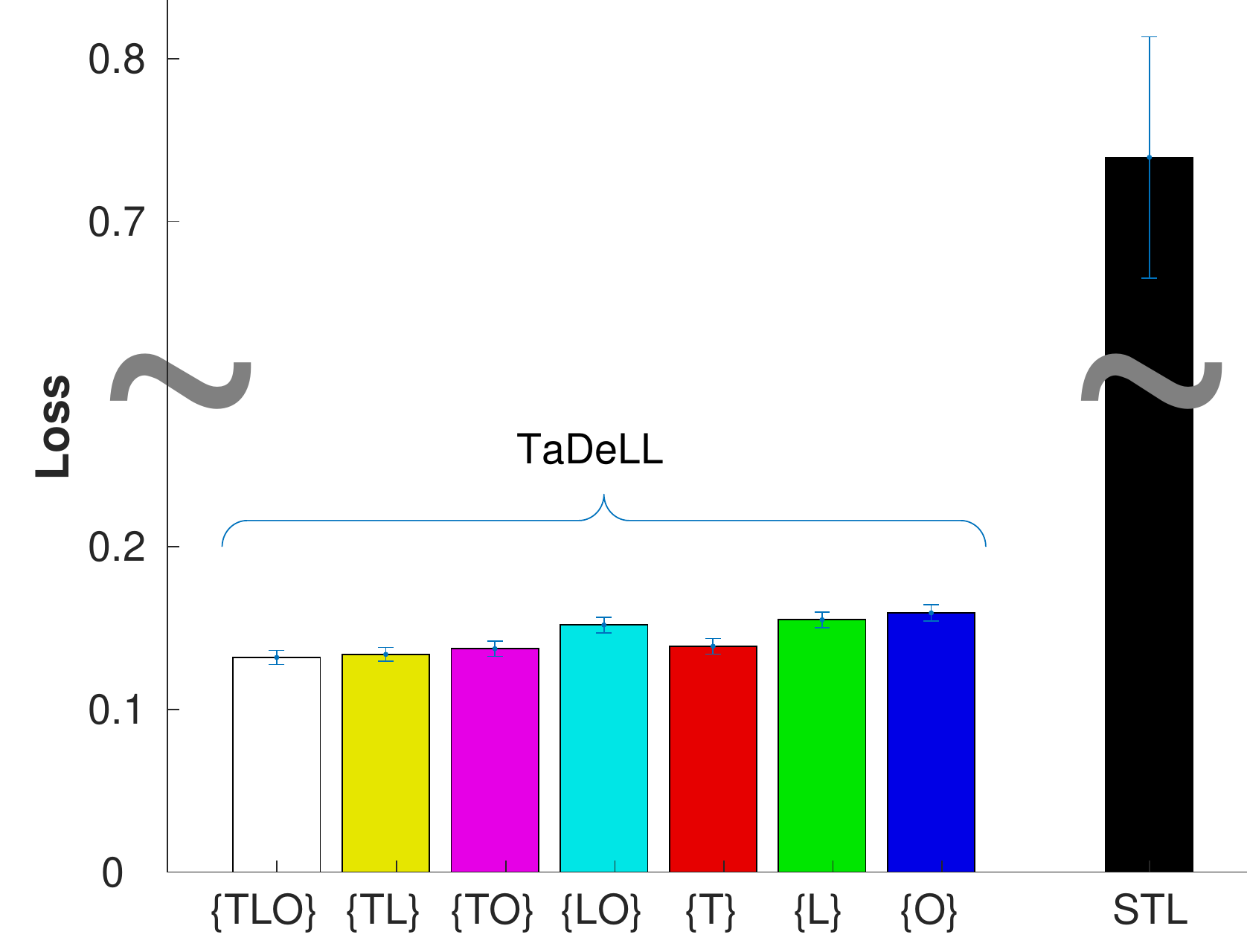}
        \caption{Robot End Effector Prediction}
        \label{fig:robot_numfeatures}
    \end{subfigure}
    	\caption{Performance using various subsets of the SM system parameters (mass $M$, damping constant $D$, and spring constant $K$) and Robot system parameters (twist $T$, link length $L$, and offset $O$) as the task descriptors.} %
\end{figure}

To address these questions, we conducted additional experiments on the Spring-Mass (SM) system and robot end effector problem, using various subsets of the task descriptor features when learning the coupled dictionaries. Figure \ref{fig:sm_numfeatures} shows how the number and selection of parameters affects performance on the SM domain.  We evaluated jumpstart performance when using all possible subsets of the system parameters as the task descriptor features.  These subsets of the SM system parameters (mass $M$, damping constant $D$, and spring constant $K$) are shown along the horizontal axis  for the task descriptors. Overall, the results show that the learner performs better when using larger subsets of the system parameters as the task descriptors.

The robot task has 24 descriptors consisting of a twist, link offset, and link length for each joint. We group the subset of features describing twist, offset, and length together and examine removing different subsets. Figure \ref{fig:robot_numfeatures} show that twist is more important than the other features and again the inclusion of more features improves performance. 

\subsection{Computational Efficiency}\label{computation}

\begin{figure}[t!]
    \centering
     \vspace{0pt}
     \includegraphics[clip,trim={97pt 245pt 132pt 262pt},width=.6\textwidth]{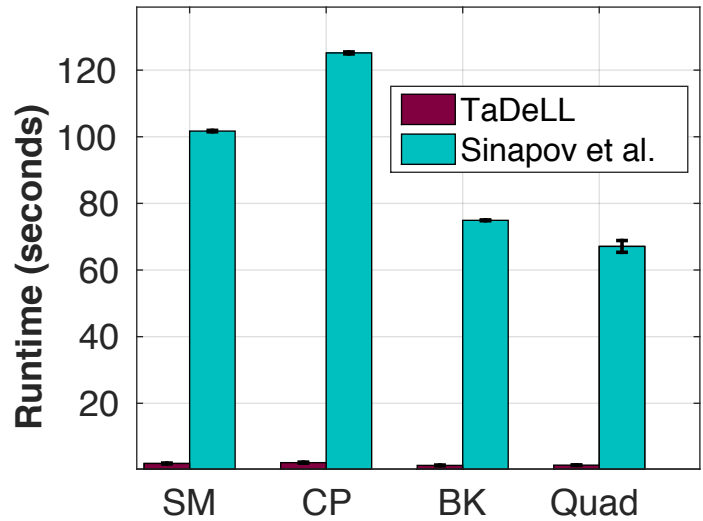}
    \vspace{-0.25em}
    \caption{Runtime comparison.\\~}
    \label{fig:time}
    
\end{figure}

We compared the average per-task runtime of our approach to that of Sinapov~\etal~\citeyear{Sinapov2015}, the most closely related method to our approach. Since Sinapov~\etal's method 
requires training transferability predictors between all pairs of tasks, its total runtime grows quadratically with the number of tasks.  
In comparison, our online algorithm is highly efficient.  
As shown in Section~\ref{sect:computationComplexity}, the per-update cost of TaDeLL is $O\left( k^2 (d+m)^3 + \xi(d,n_t) \right)$. Note that this per-update cost is independent of the number of tasks $T$, giving TaDeLL a total runtime that scales linearly in the number of tasks.

Figure \ref{fig:time} shows the per-task runtime for each algorithm based on a set of 40 tasks, as evaluated on an Intel Core I7-4700HQ CPU.  
TaDeLL samples tasks randomly with replacement and terminates once every task has been seen. For Sinapov \etal, we used 10 PG iterations for calculating the warm start, ensuring fair comparison between the methods. %
These results show a substantial reduction in computational time for TaDeLL: two orders of magnitude over the 40 tasks.

\subsection{Performance for Various Numbers of Tasks}

\begin{figure}[!htbp]
    \centering
    \begin{subfigure}[b]{3in}
        \includegraphics[clip,trim={100pt 220pt 100pt 250pt},height=2.0in]{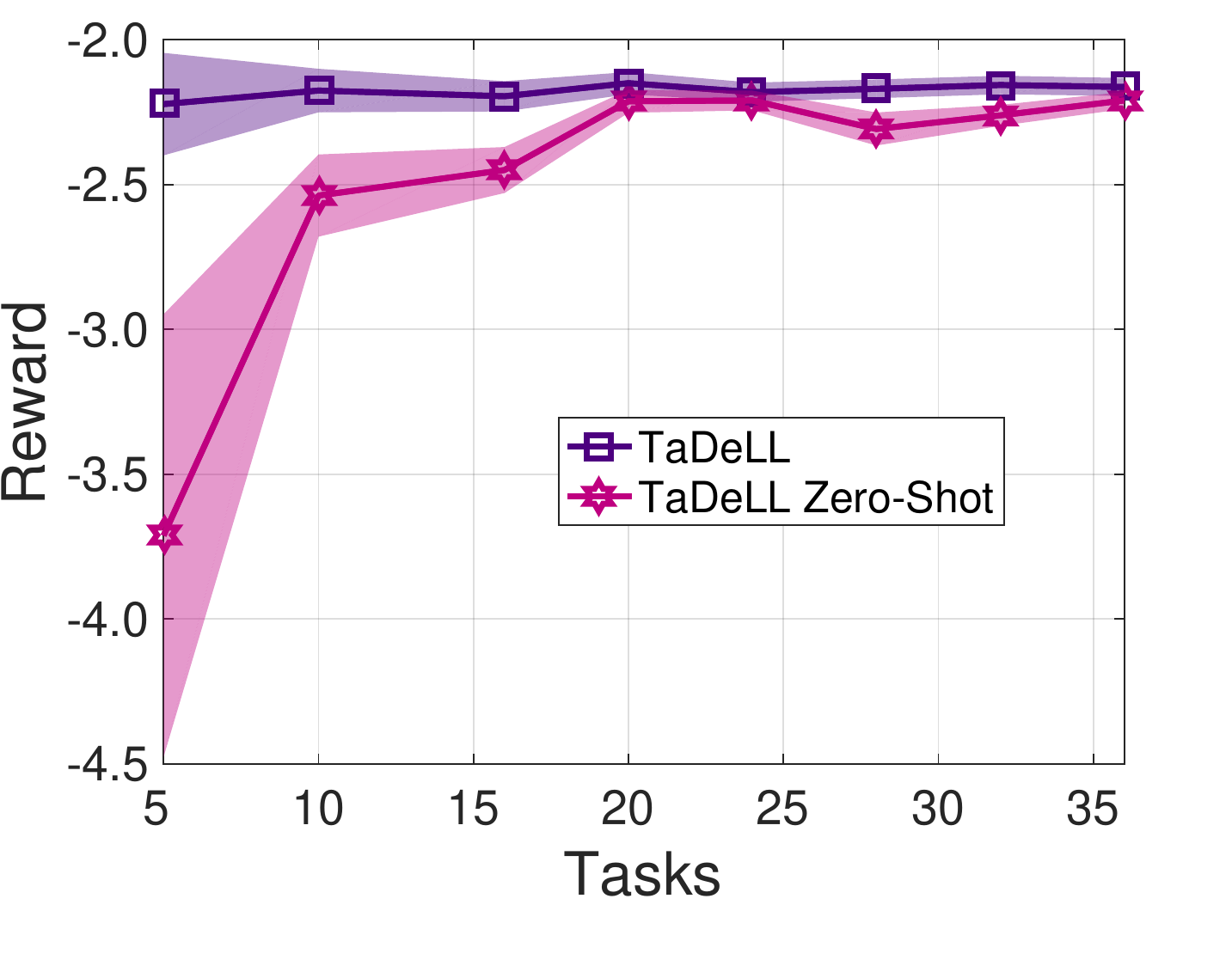}
        \caption{Spring-Mass RL}
        \end{subfigure}~
    \begin{subfigure}[b]{3in}
    	\includegraphics[clip,trim={100pt 220pt 100pt 250pt},height=2.0in]{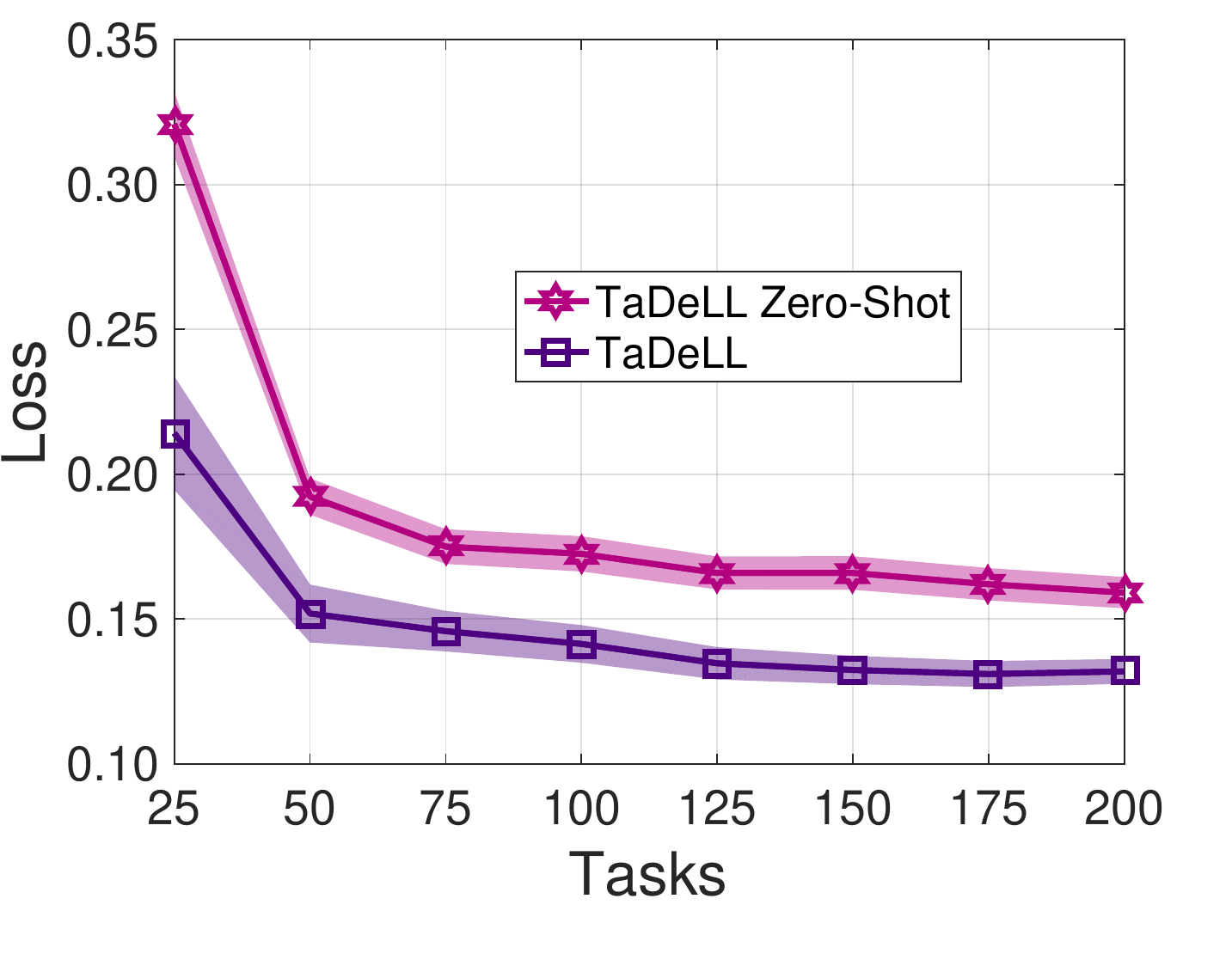}
        \caption{Robot End Effector Prediction}
    \end{subfigure}
    \caption{Zero-shot performance as a function of the number of tasks used to train the dictionary. As more tasks are used, the performance of zero-shot transfer improves. 
        } \label{fig:tasknum}
\end{figure}

Although we have shown in Section~\ref{sect:convergence} that the learned dictionaries become more stable as the system learns more tasks, we cannot currently guarantee that this will improve the performance of zero-shot transfer.  To  evaluate the effect of the number of tasks on zero-shot performance, we conducted an additional set of experiments on both the Simple-Mass domain and the robot end effector prediction domain.  Our results, shown in Figure \ref{fig:tasknum}, reveal that zero-shot performance does indeed improve as the dictionaries are trained over more tasks.  This improvement is most stable and rapid in an MTL setting, since the optimization over all dictionaries and task policies is run to convergence, but TaDeLL also shows clear improvement in zero-shot performance as $T_{\mathit{max}}$ increases.  Since zero-shot transfer involves only the learned coupled dictionaries, we can conclude that the quality of these dictionaries for zero-shot transfer improves as the system learns more tasks.

\section{Conclusion} \label{sect:Conclusion}

This article demonstrated that incorporating high-level task descriptors into lifelong learning both improves learning performance and also enables zero-shot transfer to new tasks.  The mechanism of using a coupled dictionary to connect the task descriptors with the learned models is relatively straightforward, yet highly effective in practice and has connections to\comment{co-training and} mutual coherence in sparse coding.  Most critically, it provides a fast and simple mechanism to predict the model or policy for a new task via zero-shot learning, given only its high level task descriptor. This approach is general and can handle multiple learning paradigms, including classification, regression, and RL tasks. Experiments demonstrate that our approach outperforms the state of the art and requires substantially less computational time than competing methods.

This ability to rapidly bootstrap models (or policies) for new tasks is critical to the development of lifelong learning systems that will be deployed for extended periods in real environments and tasked with handling a variety of tasks.  High-level descriptions provide an effective way for humans %
to communicate and to instruct each other.  The description need not come from another agent; humans often read instructions and then complete a novel task quite effectively.  Enabling lifelong learning systems to similarly take advantage of these high-level descriptions provides an effective step toward their practical effectiveness.  As shown in our experiments with warm-start learning from the zero-shot predicted policy, these task descriptors can also be combined with training data on the new task in a hybrid approach.

Despite TaDeLL's strong performance, defining what constitutes an effective task descriptor for a group of related tasks remains an open question.  In our framework, task descriptors are given, typically as fundamental descriptions of the system.  The representation we use for the task descriptors, a feature vector, is also relatively simple.  One interesting direction for future work is to develop methods for integrating more complex task descriptors into MTL or lifelong learning.  These more sophisticated mechanism could include natural language descriptions, step-by-step instructions, or logical relationships.  Such an advance would likely involve moving beyond the linear framework used in TaDeLL, but would constitute an important step toward enabling more practical use of high-level task descriptors in lifelong learning.

\section*{Acknowledgments}

This research was supported by ONR grant \#N00014-11-1-0139, AFRL grant \#FA8750-14-1-0069, and AFRL grant \#FA8750-16-1-0109. We would like
to thank the anonymous reviewers of the conference version of this paper for their helpful feedback.

\vskip 0.2in
\bibliography{AI_Lifelong_Learning}

\begin{thebibliography}{}

\bibitem[\protect\BCAY{Ando\ \BBA\ Zhang}{Ando\ \BBA\
  Zhang}{2005}]{ando2005framework}
Ando, R.~K.\BBACOMMA\  \BBA\ Zhang, T. \BBOP2005\BBCP.
\newblock \BBOQ A framework for learning predictive structures from multiple
  tasks and unlabeled data\BBCQ\
\newblock {\Bem The Journal of Machine Learning Research}, {\Bem 6},
  1817--1853.

\bibitem[\protect\BCAY{Bakker\ \BBA\ Heskes}{Bakker\ \BBA\
  Heskes}{2003}]{Bakker2003}
Bakker, B.\BBACOMMA\  \BBA\ Heskes, T. \BBOP2003\BBCP.
\newblock \BBOQ {Task clustering and gating for Bayesian multitask
  learning}\BBCQ\
\newblock {\Bem The Journal of Machine Learning Research}, {\Bem 4}, 83--99.

\bibitem[\protect\BCAY{Baxter}{Baxter}{2000}]{baxter2000model}
Baxter, J. \BBOP2000\BBCP.
\newblock \BBOQ A model of inductive bias learning\BBCQ\
\newblock {\Bem The Journal of Artificial Intelligence Research}, {\Bem 12},
  149--198.

\bibitem[\protect\BCAY{Bickel, Sawade,\ \BBA\ Scheffer}{Bickel
  et~al.}{2009}]{Bickel2008}
Bickel, S., Sawade, C., \BBA\ Scheffer, T. \BBOP2009\BBCP.
\newblock \BBOQ {Transfer learning by distribution matching for targeted
  advertising}\BBCQ\
\newblock {\Bem Advances in Neural Information Processing Systems}, 145--152.

\bibitem[\protect\BCAY{Bonilla, Agakov,\ \BBA\ Williams}{Bonilla
  et~al.}{2007}]{bonilla2007kernel}
Bonilla, E.~V., Agakov, F.~V., \BBA\ Williams, C. \BBOP2007\BBCP.
\newblock \BBOQ Kernel multi-task learning using task-specific features\BBCQ\
\newblock {\Bem In Proceedings of the International Conference on Artificial
  Intelligence and Statistics}, 43--50.

\bibitem[\protect\BCAY{Bou~Ammar, Eaton, Luna,\ \BBA\ Ruvolo}{Bou~Ammar
  et~al.}{2015}]{bouammar2015autonomous}
Bou~Ammar, H., Eaton, E., Luna, J.~M., \BBA\ Ruvolo, P. \BBOP2015\BBCP.
\newblock \BBOQ Autonomous cross-domain knowledge transfer in lifelong policy
  gradient reinforcement learning\BBCQ\
\newblock {\Bem In Proceedings of the International Joint Conference on
  Artificial Intelligence}.

\bibitem[\protect\BCAY{Bou~Ammar, Eaton,\ \BBA\ Ruvolo}{Bou~Ammar
  et~al.}{2014}]{Ammar2014a}
Bou~Ammar, H., Eaton, E., \BBA\ Ruvolo, P. \BBOP2014\BBCP.
\newblock \BBOQ {Online multi-task learning for policy gradient methods}\BBCQ\
\newblock {\Bem In Proceedings of the International Conference on Machine
  Learning}.

\bibitem[\protect\BCAY{{Bou Ammar}, Eaton, Ruvolo,\ \BBA\ Taylor}{{Bou Ammar}
  et~al.}{2015}]{BouAmmar2015Unsupervised}
{Bou Ammar}, H., Eaton, E., Ruvolo, P., \BBA\ Taylor, M.~E. \BBOP2015\BBCP.
\newblock \BBOQ Unsupervised cross-domain transfer in policy gradient
  reinforcement learning via manifold alignment\BBCQ\
\newblock In {\Bem Proceedings of the AAAI Conference on Artificial
  Intelligence (AAAI-15)}.

\bibitem[\protect\BCAY{Bouabdallah\ \BBA\ Siegwart}{Bouabdallah\ \BBA\
  Siegwart}{2005}]{bouabdallah2005backstepping}
Bouabdallah, S.\BBACOMMA\  \BBA\ Siegwart, R. \BBOP2005\BBCP.
\newblock \BBOQ Backstepping and sliding-mode techniques applied to an indoor
  micro quadrotor\BBCQ\
\newblock {\Bem In Proceedings of the 2005 IEEE International Conference on
  Robotics and Automation.}, 2247--2252.

\bibitem[\protect\BCAY{Caruana}{Caruana}{1997}]{Caruana1997}
Caruana, R. \BBOP1997\BBCP.
\newblock \BBOQ {Multitask Learning}\BBCQ\
\newblock {\Bem Machine Learning}, {\Bem 28}, 41--75.

\bibitem[\protect\BCAY{Cavallanti, Cesa-Bianchi,\ \BBA\ Gentile}{Cavallanti
  et~al.}{2010}]{cavallanti2010linear}
Cavallanti, G., Cesa-Bianchi, N., \BBA\ Gentile, C. \BBOP2010\BBCP.
\newblock \BBOQ Linear algorithms for online multitask classification\BBCQ\
\newblock {\Bem The Journal of Machine Learning Research}, {\Bem 11},
  2901--2934.

\bibitem[\protect\BCAY{Corke}{Corke}{2011}]{Corkella}
Corke, P.~I. \BBOP2011\BBCP.
\newblock \BBOQ Autonomous cross-domain knowledge transfer in lifelong policy
  gradient reinforcement learning\BBCQ\
\newblock In {\Bem Robotics, Vision \& Control: Fundamental Algorithms in
  Matlab}. Springer.

\bibitem[\protect\BCAY{Dekel, Long,\ \BBA\ Singer}{Dekel
  et~al.}{2006}]{dekel2006online}
Dekel, O., Long, P.~M., \BBA\ Singer, Y. \BBOP2006\BBCP.
\newblock \BBOQ Online multitask learning\BBCQ\
\newblock In {\Bem Proceedings of the International Conference on Computational
  Learning Theory}, \BPGS\ 453--467. Springer.

\bibitem[\protect\BCAY{Donoho, Elad,\ \BBA\ Temlyakov}{Donoho
  et~al.}{2006}]{donoho2006stable}
Donoho, D.~L., Elad, M., \BBA\ Temlyakov, V.~N. \BBOP2006\BBCP.
\newblock \BBOQ Stable recovery of sparse overcomplete representations in the
  presence of noise\BBCQ\
\newblock {\Bem IEEE Transactions on Information Theory}, {\Bem 52\/}(1),
  6--18.

\bibitem[\protect\BCAY{Donoho\ \BBA\ Huo}{Donoho\ \BBA\
  Huo}{2001}]{donoho2001uncertainty}
Donoho, D.~L.\BBACOMMA\  \BBA\ Huo, X. \BBOP2001\BBCP.
\newblock \BBOQ Uncertainty principles and ideal atomic decomposition\BBCQ\
\newblock {\Bem IEEE Transactions on Information Theory}, {\Bem 47\/}(7),
  2845--2862.

\bibitem[\protect\BCAY{Evgeniou\ \BBA\ Pontil}{Evgeniou\ \BBA\
  Pontil}{2004}]{evgeniou2004regularized}
Evgeniou, T.\BBACOMMA\  \BBA\ Pontil, M. \BBOP2004\BBCP.
\newblock \BBOQ Regularized multi--task learning\BBCQ\
\newblock In {\Bem Proceedings of the International Conference on Knowledge
  Discovery and Data Mining}, \BPGS\ 109--117. ACM.

\bibitem[\protect\BCAY{Ham, Lee,\ \BBA\ Saul}{Ham
  et~al.}{2005}]{ham2005semisupervised}
Ham, J., Lee, D.~D., \BBA\ Saul, L.~K. \BBOP2005\BBCP.
\newblock \BBOQ Semisupervised alignment of manifolds\BBCQ\
\newblock In {\Bem Proceedings of International Conference on Artificial
  Intelligence and Statistics}, \BPGS\ 120--127.

\bibitem[\protect\BCAY{Huang, Socher, Manning,\ \BBA\ Ng}{Huang
  et~al.}{2012}]{Huang2012}
Huang, E.~H., Socher, R., Manning, C.~D., \BBA\ Ng, A. \BBOP2012\BBCP.
\newblock \BBOQ {Improving word representations via global context and multiple
  word prototypes}\BBCQ\
\newblock {\Bem Proceedings of the 50th Annual Meeting of the Association for
  Computational Linguistics: Long Papers-Volume 1}, 873--882.

\bibitem[\protect\BCAY{Isele, Rostami,\ \BBA\ Eaton}{Isele
  et~al.}{2016}]{isele2016task}
Isele, D., Rostami, M., \BBA\ Eaton, E. \BBOP2016\BBCP.
\newblock \BBOQ Using task features for zero-shot knowledge transfer in
  lifelong learning\BBCQ\
\newblock {\Bem In Proceedings of the International Joint Conference on
  Artificial Intelligence}.

\bibitem[\protect\BCAY{Kober\ \BBA\ Peters}{Kober\ \BBA\
  Peters}{2009}]{kober2009policy}
Kober, J.\BBACOMMA\  \BBA\ Peters, J. \BBOP2009\BBCP.
\newblock \BBOQ Policy search for motor primitives in robotics\BBCQ\
\newblock {\Bem Advances in Neural Information Processing Systems}, 849--856.

\bibitem[\protect\BCAY{Kumar\ \BBA\ Daum{\'e}}{Kumar\ \BBA\
  Daum{\'e}}{2012}]{Kumar2012}
Kumar, A.\BBACOMMA\  \BBA\ Daum{\'e}, H. \BBOP2012\BBCP.
\newblock \BBOQ {Learning task grouping and overlap in multi-task
  learning}\BBCQ\
\newblock {\Bem In Proceedings of the International Conference on Machine
  Learning}, 1383--1390.

\bibitem[\protect\BCAY{Lazaric\ \BBA\ Ghavamzadeh}{Lazaric\ \BBA\
  Ghavamzadeh}{2010}]{lazaric2010bayesian}
Lazaric, A.\BBACOMMA\  \BBA\ Ghavamzadeh, M. \BBOP2010\BBCP.
\newblock \BBOQ Bayesian multi-task reinforcement learning\BBCQ\
\newblock In {\Bem Proceedings of International Conference on Machine
  Learning}, \BPGS\ 599--606. Omnipress.

\bibitem[\protect\BCAY{Maurer, Pontil,\ \BBA\ Romera-Paredes}{Maurer
  et~al.}{2013}]{Maurer2013}
Maurer, A., Pontil, M., \BBA\ Romera-Paredes, B. \BBOP2013\BBCP.
\newblock \BBOQ {Sparse coding for multitask and transfer learning}\BBCQ\
\newblock {\Bem In Proceedings of the International Conference on Machine
  Learning}, {\Bem 28}, 343--351.

\bibitem[\protect\BCAY{Negahban, Yu, Wainwright,\ \BBA\ Ravikumar}{Negahban
  et~al.}{2009}]{negahban2009unified}
Negahban, S., Yu, B., Wainwright, M., \BBA\ Ravikumar, P. \BBOP2009\BBCP.
\newblock \BBOQ A unified framework for high-dimensional analysis of $ m
  $-estimators with decomposable regularizers\BBCQ\
\newblock In {\Bem Advances in Neural Information Processing Systems}, \BPGS\
  1348--1356.

\bibitem[\protect\BCAY{Oyen\ \BBA\ Lane}{Oyen\ \BBA\
  Lane}{2012}]{oyen2012leveraging}
Oyen, D.\BBACOMMA\  \BBA\ Lane, T. \BBOP2012\BBCP.
\newblock \BBOQ Leveraging domain knowledge in multitask {B}ayesian network
  structure learning\BBCQ\
\newblock In {\Bem Proceedings of the AAAI Conference on Artificial
  Intelligence}.

\bibitem[\protect\BCAY{Palatucci, Hinton, Pomerleau,\ \BBA\ Mitchell}{Palatucci
  et~al.}{2009}]{Palatucci2009}
Palatucci, M., Hinton, G., Pomerleau, D., \BBA\ Mitchell, T.~M. \BBOP2009\BBCP.
\newblock \BBOQ {Zero-shot learning with semantic output codes}\BBCQ\
\newblock {\Bem Advances in Neural Information Processing Systems}.

\bibitem[\protect\BCAY{Pan\ \BBA\ Yang}{Pan\ \BBA\ Yang}{2010}]{Pan2010a}
Pan, S.~J.\BBACOMMA\  \BBA\ Yang, Q. \BBOP2010\BBCP.
\newblock \BBOQ {A survey on transfer learning}\BBCQ\
\newblock {\Bem IEEE Transactions on Knowledge and Data Engineering}, {\Bem
  22\/}(10).

\bibitem[\protect\BCAY{Pennington, Socher,\ \BBA\ Manning}{Pennington
  et~al.}{2014}]{pennington2014glove}
Pennington, J., Socher, R., \BBA\ Manning, C.~D. \BBOP2014\BBCP.
\newblock \BBOQ Glove: Global vectors for word representation\BBCQ\
\newblock {\Bem Proceedings of the Empiricial Methods in Natural Language
  Processing (EMNLP 2014)}, {\Bem 12}, 1532--1543.

\bibitem[\protect\BCAY{Peters\ \BBA\ Schaal}{Peters\ \BBA\
  Schaal}{2008}]{peters2008natural}
Peters, J.\BBACOMMA\  \BBA\ Schaal, S. \BBOP2008\BBCP.
\newblock \BBOQ Natural actor-critic\BBCQ\
\newblock {\Bem Neurocomputing}, {\Bem 71\/}(7), 1180--1190.

\bibitem[\protect\BCAY{Romera-Paredes\ \BBA\ Torr}{Romera-Paredes\ \BBA\
  Torr}{2015}]{romera2015embarrassingly}
Romera-Paredes, B.\BBACOMMA\  \BBA\ Torr, P. H.~S. \BBOP2015\BBCP.
\newblock \BBOQ An embarrassingly simple approach to zero-shot learning\BBCQ\
\newblock {\Bem Proceedings of International Conference on Machine Learning},
  2152--2161.

\bibitem[\protect\BCAY{Ruvolo\ \BBA\ Eaton}{Ruvolo\ \BBA\
  Eaton}{2013}]{Ruvolo2013}
Ruvolo, P.\BBACOMMA\  \BBA\ Eaton, E. \BBOP2013\BBCP.
\newblock \BBOQ {ELLA: An efficient lifelong learning algorithm}\BBCQ\
\newblock {\Bem Proceedings of the International Conference on Machine
  Learning}, {\Bem 28}, 507--515.

\bibitem[\protect\BCAY{Saha, Rai, Venkatasubramanian,\ \BBA\ Daume}{Saha
  et~al.}{2011}]{saha2011online}
Saha, A., Rai, P., Venkatasubramanian, S., \BBA\ Daume, H. \BBOP2011\BBCP.
\newblock \BBOQ Online learning of multiple tasks and their relationships\BBCQ\
\newblock {\Bem Proceedings of International Conference on Artificial
  Intelligence and Statistics}, 643--651.

\bibitem[\protect\BCAY{Sinapov, Narvekar, Leonetti,\ \BBA\ Stone}{Sinapov
  et~al.}{2015}]{Sinapov2015}
Sinapov, J., Narvekar, S., Leonetti, M., \BBA\ Stone, P. \BBOP2015\BBCP.
\newblock \BBOQ {Learning inter-task transferability in the absence of target
  task samples}\BBCQ\
\newblock {\Bem Proceedings of the 14th International Conference on Autonomous
  Agents and Multiagent Systems}.

\bibitem[\protect\BCAY{Socher, Ganjoo, Manning,\ \BBA\ Ng}{Socher
  et~al.}{2013}]{Socher2013}
Socher, R., Ganjoo, M., Manning, C.~D., \BBA\ Ng, A.~Y. \BBOP2013\BBCP.
\newblock \BBOQ {Zero-shot learning through cross-modal transfer}\BBCQ\
\newblock {\Bem Advances in Neural Information Processing Systems}, 935--943.

\bibitem[\protect\BCAY{Sutton, McAllester, Singh,\ \BBA\ Mansour}{Sutton
  et~al.}{1999}]{sutton1999policy}
Sutton, R.~S., McAllester, D.~A., Singh, S.~P., \BBA\ Mansour, Y.
  \BBOP1999\BBCP.
\newblock \BBOQ Policy gradient methods for reinforcement learning with
  function approximation\BBCQ\
\newblock {\Bem Advances in Neural Information Processing Systems}, {\Bem 99},
  1057--1063.

\bibitem[\protect\BCAY{Taylor\ \BBA\ Stone}{Taylor\ \BBA\
  Stone}{2009}]{taylor2009transfer}
Taylor, M.~E.\BBACOMMA\  \BBA\ Stone, P. \BBOP2009\BBCP.
\newblock \BBOQ Transfer learning for reinforcement learning domains: A
  survey\BBCQ\
\newblock {\Bem The Journal of Machine Learning Research}, {\Bem 10},
  1633--1685.

\bibitem[\protect\BCAY{Taylor, Stone,\ \BBA\ Liu}{Taylor
  et~al.}{2007}]{taylor2007transfer}
Taylor, M.~E., Stone, P., \BBA\ Liu, Y. \BBOP2007\BBCP.
\newblock \BBOQ Transfer learning via inter-task mappings for temporal
  difference learning\BBCQ\
\newblock {\Bem The Journal of Machine Learning Research}, {\Bem 8\/}(Sep),
  2125--2167.

\bibitem[\protect\BCAY{Thrun}{Thrun}{1996}]{Thrun1996}
Thrun, S. \BBOP1996\BBCP.
\newblock \BBOQ Is learning the n-th thing any easier than learning the
  first?\BBCQ\
\newblock {\Bem Advances in Neural Information Processing Systems}, 640--646.

\bibitem[\protect\BCAY{Wang\ \BBA\ Mahadevan}{Wang\ \BBA\
  Mahadevan}{2009}]{wang2009general}
Wang, C.\BBACOMMA\  \BBA\ Mahadevan, S. \BBOP2009\BBCP.
\newblock \BBOQ A general framework for manifold alignment\BBCQ\
\newblock In {\Bem Proceedings of the AAAI Conference on Artificial
  Intelligence}.

\bibitem[\protect\BCAY{Williams}{Williams}{1992}]{williams1992simple}
Williams, R.~J. \BBOP1992\BBCP.
\newblock \BBOQ Simple statistical gradient-following algorithms for
  connectionist reinforcement learning\BBCQ\
\newblock {\Bem Machine Learning}, {\Bem 8\/}(3-4), 229--256.

\bibitem[\protect\BCAY{Wilson, Fern, Ray,\ \BBA\ Tadepalli}{Wilson
  et~al.}{2007}]{wilson2007multi}
Wilson, A., Fern, A., Ray, S., \BBA\ Tadepalli, P. \BBOP2007\BBCP.
\newblock \BBOQ Multi-task reinforcement learning: a hierarchical bayesian
  approach\BBCQ\
\newblock In {\Bem Proceedings of the International Conference on Machine
  Learning}, \BPGS\ 1015--1022. ACM.

\bibitem[\protect\BCAY{Xu, Hospedales,\ \BBA\ Gong}{Xu
  et~al.}{2016}]{xu2016multi}
Xu, X., Hospedales, T.~M., \BBA\ Gong, S. \BBOP2016\BBCP.
\newblock \BBOQ Multi-task zero-shot action recognition with prioritised data
  augmentation\BBCQ\
\newblock In {\Bem Proceedings of the European Conference on Computer Vision},
  \BPGS\ 343--359. Springer.

\bibitem[\protect\BCAY{Yang, Wright, Huang,\ \BBA\ Ma}{Yang
  et~al.}{2010}]{yang2010image}
Yang, J., Wright, J., Huang, T.~S., \BBA\ Ma, Y. \BBOP2010\BBCP.
\newblock \BBOQ Image super-resolution via sparse representation\BBCQ\
\newblock {\Bem IEEE Transactions on Image Processing}, {\Bem 19\/}(11),
  2861--2873.

\bibitem[\protect\BCAY{Yu, Wu, Yang, Tian, Luo,\ \BBA\ Zhuang}{Yu
  et~al.}{2014}]{yu2014discriminative}
Yu, Z., Wu, F., Yang, Y., Tian, Q., Luo, J., \BBA\ Zhuang, Y. \BBOP2014\BBCP.
\newblock \BBOQ Discriminative coupled dictionary hashing for fast cross-media
  retrieval\BBCQ\
\newblock {\Bem Proceedings of the 37th International ACM SIGIR Conference on
  Research \& Development in Information Retrieval}, 395--404.

\bibitem[\protect\BCAY{Zhong\ \BBA\ Kwok}{Zhong\ \BBA\ Kwok}{2012}]{Zhong2012a}
Zhong, L.~W.\BBACOMMA\  \BBA\ Kwok, J.~T. \BBOP2012\BBCP.
\newblock \BBOQ {Convex multitask learning with flexible task clusters}\BBCQ\
\newblock {\Bem Proceedings of the International Conference on Machine
  Learning}, {\Bem 1}, 49--56.

\bibitem[\protect\BCAY{Zhuang, Wang, Wu, Zhang,\ \BBA\ Lu}{Zhuang
  et~al.}{2013}]{zhuang2013supervised}
Zhuang, Y.~T., Wang, Y.~F., Wu, F., Zhang, Y., \BBA\ Lu, W.~M. \BBOP2013\BBCP.
\newblock \BBOQ Supervised coupled dictionary learning with group structures
  for multi-modal retrieval\BBCQ\
\newblock {\Bem Proceedings of the Twenty-Seventh AAAI Conference on Artificial
  Intelligence}.

\end{thebibliography}
\bibliographystyle{theapa}

\end{document}